\RequirePackage{fix-cm}
\documentclass[twocolumn]{svjour3}          % twocolumn
\smartqed  % flush right qed marks, e.g. at end of proof
\usepackage{graphicx}
\usepackage[table]{xcolor}
\usepackage{xspace}

\newcommand{\eg}{\emph{e.g.}\xspace}
\newcommand{\ie}{\emph{i.e.}\xspace}

\newcommand{\etc}{\emph{etc.}\xspace}
\newcommand{\argmax}{\operatornamewithlimits{argmax}}

\makeatletter
\newcommand\footnoteref[1]{\protected@xdef\@thefnmark{\ref{#1}}\@footnotemark}
\makeatother

\usepackage{color}

\usepackage{multirow} 
\usepackage{natbib}
\usepackage{amsmath}
\usepackage{amssymb}
\usepackage[breaklinks=true]{hyperref}
\usepackage{verbatim}

\usepackage{array}
\newcolumntype{L}[1]{>{\raggedright\let\newline\\\arraybackslash\hspace{0pt}}m{#1}}
\newcolumntype{C}[1]{>{\centering\let\newline\\\arraybackslash\hspace{0pt}}m{#1}}
\newcolumntype{R}[1]{>{\raggedleft\let\newline\\\arraybackslash\hspace{0pt}}m{#1}}

\usepackage{hyphenat}
\let\oldhref\href
\renewcommand{\href}[2]{\oldhref{#1}{\hbox{#2}}}

\begin{document}\sloppy

%\title{Solving Visual Madlibs with Multiple Cues
\title{
%\red{
Combining Multiple Cues for \\Visual Madlibs Question Answering
}

%\titlerunning{Short form of title}  % if too long for running head

\author{Tatiana Tommasi         \and
        Arun Mallya         \and  
        Bryan  Plummer         \and  
        Svetlana  Lazebnik         \and 
        Alexander C. Berg         \and
        Tamara L. Berg
}


\institute{T. Tommasi \at
			  Dept. of Computer Control and Management Engineering
              University of Rome, La Sapienza, Italy\\
              \email{tommasi@dis.uniroma1.it}           %  \\
           \and
           A. Mallya \at
           University of Illinois at Urbana Champaign, IL, USA
           \and
           B. A. Plummer \at
           University of Illinois at Urbana Champaign, IL, USA
           \and
           S. Lazebnik\at
           University of Illinois at Urbana Champaign, IL, USA
           \and
           A. C. Berg\at
           University of North Carolina at Chapel Hill, NC, USA
           \and
           T. L. Berg\at
           University of North Carolina at Chapel Hill, NC, USA
}

\date{Received: date / Accepted: date}

\maketitle

\begin{abstract}
%!TEX root = main.tex

This paper presents an approach for answering fill-in-the-blank multiple choice questions from the Visual Madlibs dataset. Instead of generic and commonly used representations trained on the ImageNet classification task, our approach employs a combination of networks trained for specialized tasks such as scene recognition, person activity classification, and attribute prediction. We also present a method for localizing phrases from candidate answers in order to provide spatial support for feature extraction. We map each of these features, together with candidate answers, to a joint embedding space through normalized canonical correlation analysis (nCCA). Finally, we solve an optimization problem to learn to combine scores from nCCA models trained on multiple cues to select the best answer. Extensive experimental results show a significant improvement over the previous state of the art and confirm that answering questions from a wide range of types benefits from examining a variety of image cues and carefully choosing the spatial support for feature extraction.

\keywords{Visual Question Answering \and Cue Integration \and Region Phrase Correspondence \and Computer Vision \and Language}
\end{abstract}

\section{Introduction}
\label{sec:introduction}

For any artificially intelligent agent that can live in the physical world, interacting with the world and communicating with humans are essential abilities. To acquire these abilities, we need to train agents on open-ended tasks that  involve visual analysis and language understanding. Visual Question Answering (VQA) \citep{VQA} has recently been proposed as such a task.  In VQA, language understanding is necessary 
to determine the intent of a question and generate or evaluate multiple putative answers, 
while visual analysis focuses on learning to extract useful information from the images. 
Even when the question has a pre-determined form, % and no semantic language interpretation is required, 
the answer strongly depends on the visual information which might be derived from either the whole image 
or from some specific image region. 
Moreover, specialized knowledge beyond the available image pixel content might be necessary.
For instance, consider a simple question about the position of an object: the answer could 
involve the overall scene (\eg, it is in the kitchen), other reference objects (\eg, it is on the table), 
their appearance (\eg, it is against the blue wall), details about people (\eg, it is in the girl's hand), 
activities (\eg, it is floating in water) or even understanding of time and causality (\eg, it is falling 
and about to land on the ground).

\begin{figure*}[t!]
\begin{center}
    \includegraphics[width=0.98\linewidth]{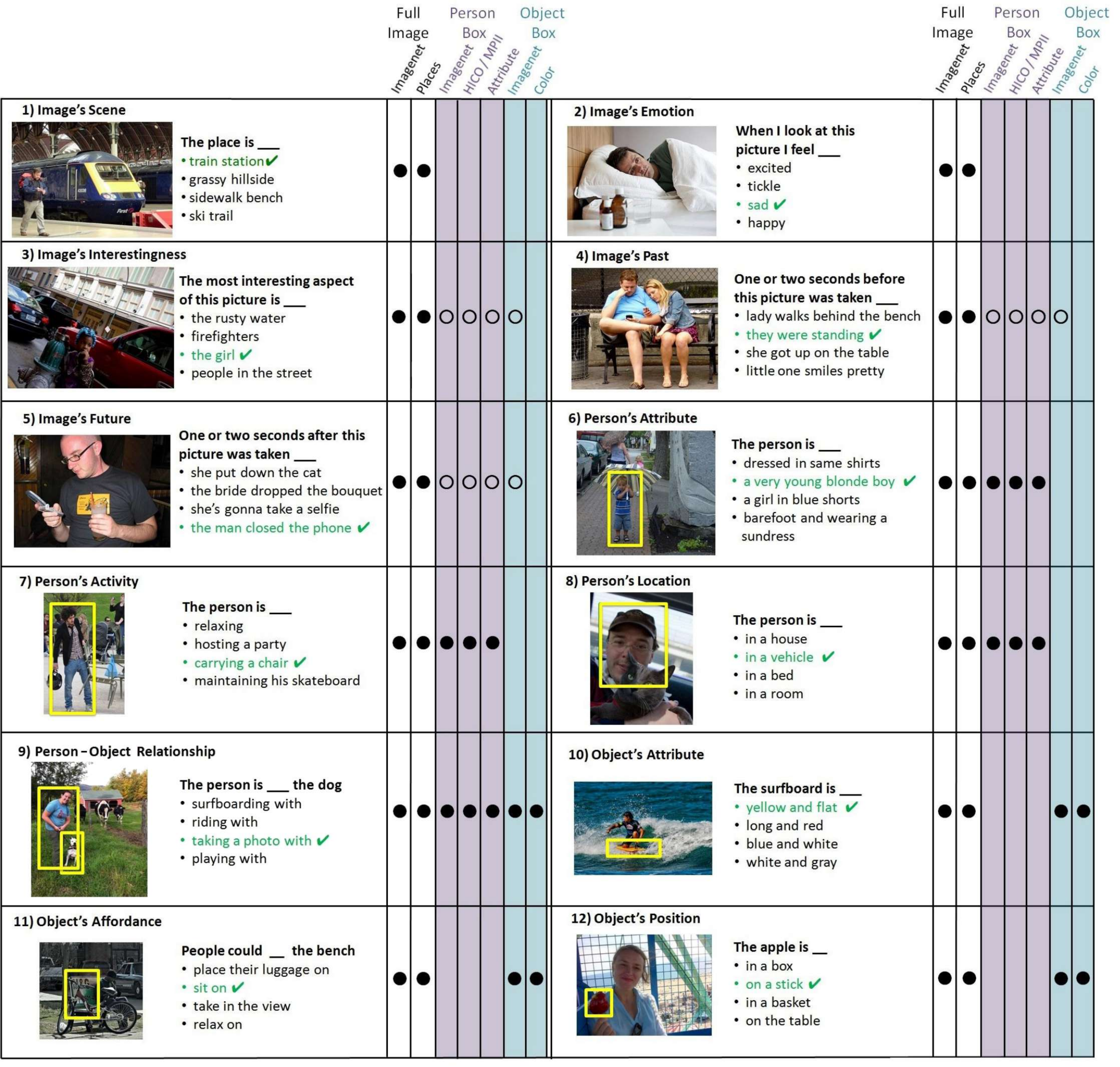} 
\caption{The Visual Madlibs dataset consists of 12 types of questions with fixed prompts, each concerned with the entire image (types 1-5), a specified person (types 6-9), or a specified object (types 9-12). For types 6-12, ground truth boxes of specified entities are provided as part of the question and are shown in yellow. Each question comes with four candidate answers, and only one (colored green, with a tick) is considered to be correct. To answer these varied questions, we use features computed on the whole image (ImageNet, Places), on person boxes (ImageNet, HICO/MPII Action, Attribute) and on object boxes (ImageNet, Color). Details of the individual cues are given in Section \ref{sec:specific_cnn_models}. For each question type, circles mark the cues that are used by our final combination method. White circles indicate that the respective cues were computed on automatically selected person and object boxes, as no ground truth boxes were provided as part of the question. All the examples here come from the Hard question-answering setting (see Section \ref{sec:related}).}

\label{fig:table_overview}
\end{center}
\vspace{-4mm}
\end{figure*}
\begin{figure*}[t!]
    \includegraphics[width=\linewidth]{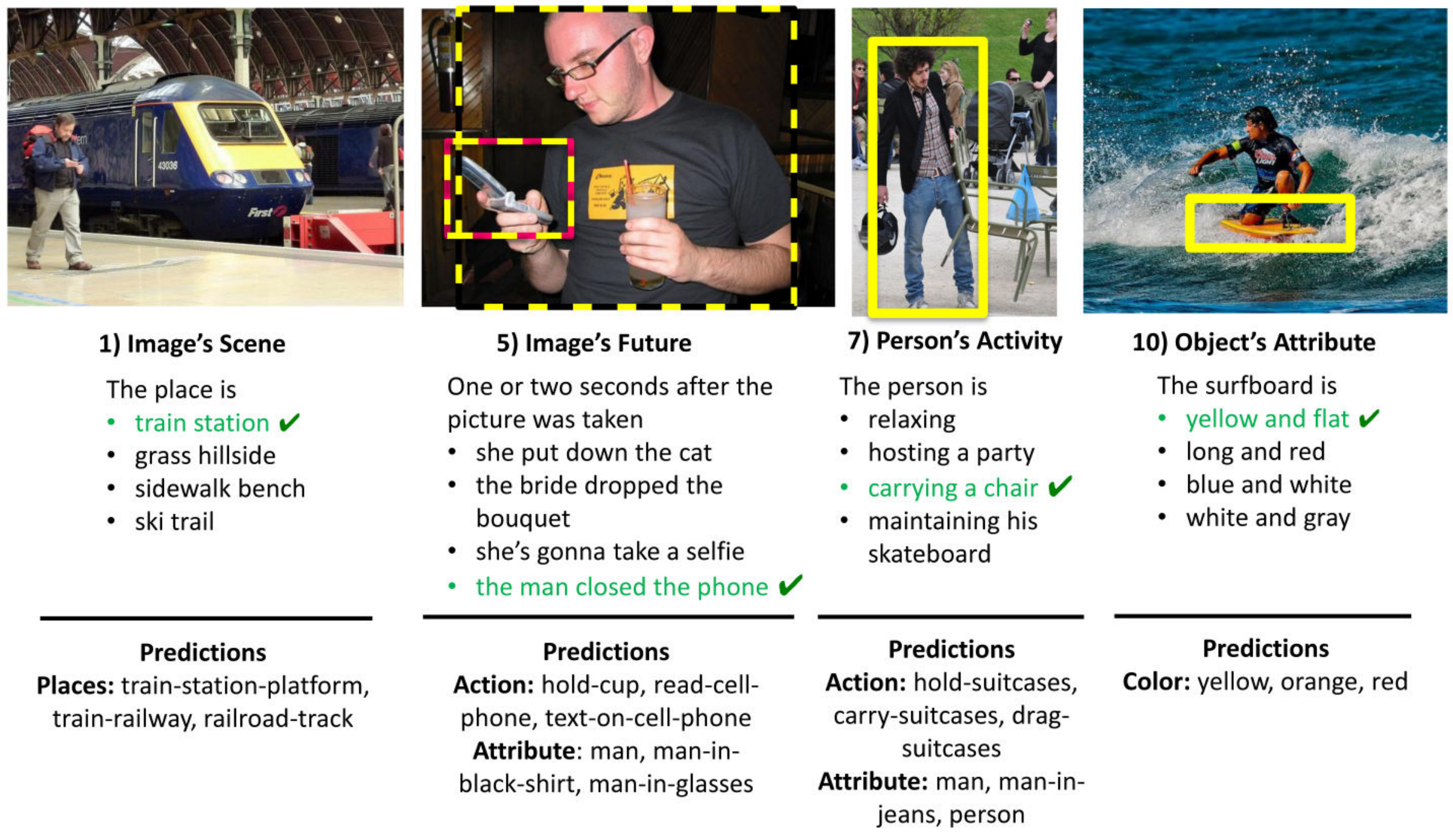}
	\caption{Examples of four questions correctly answered by our system, along with intermediate predictions from our cue-specific deep networks. For each question, three top-scoring labels from the relevant networks are shown along the bottom. For the Future question, our method automatically selects the person and phone bounding boxes (shown with dashed lines), while for the Person's Activity and Object's Attribute questions, bounding boxes are provided (solid yellow).}
    \label{fig:overview}
\end{figure*}

To date, a number of diverse solutions for VQA have been proposed, as surveyed in Section \ref{sec:related}. 
An essential component of these methods consists of extracting features from images and questions, which are 
then combined by different algorithms to produce or select the correct answer. A majority of the work has 
focused on improving such algorithms, 
while the effect of input features has been ignored: all the existing approaches use a single image representation 
computed by a deep Convolutional Neural Network (CNN), e.g. VGG-Net~\citep{simonyanVGG}, GoogLeNet~\citep{googlenet}
or ResNet \citep{resnet}
trained on the ImageNet dataset \citep{ILSVRC15}.  
While these are with no doubt powerful representations for a plethora of tasks, it is hard 
to believe that a generic feature trained on a limited number of object classes can have sufficiently broad coverage 
and fine-grained discriminative power needed to answer a wide variety of visual questions.
We believe that to truly understand an image and answer questions about it, it is necessary to leverage a rich set 
of visual cues from different sources, and to consider both global and local information. Driven by this belief, in this 
paper, we propose methods to represent the images with multiple predicted cues and introduce a learning approach to 
combine them for solving multiple-choice fill-in-the-blank style questions from the Visual Madlibs 
dataset~\citep{VisualMadlibs}.

The Visual Madlibs dataset consists of twelve different types of targeted image descriptions that have been collected by using 
fill-in-the-blank templates. For every description type, a multiple choice answering task has been defined where the sentence 
prompt takes on the role of a question, while four possible sentence completions are provided as answer options with only one 
considered to be correct (or most appropriate). Examples are shown in Figure~\ref{fig:table_overview}.
Types 1-5 are based on high-level content of the whole image, namely predicting the scene, the emotion evoked 
by the image, likely past and future events, and the most interesting aspects of the image. 
Types 6-8 are based on characteristics of a specified human subject, 9 is based on the interaction of a specified human 
and a specified object, while 10-12 are based on characteristics of a specified object. The person or object boxes that 
question types 6-12 focus on are provided as part of the question. By choosing this setting for VQA, we simplify the overall problem as we do not have to infer the question type from provided text and we can 
% we are free from latent question-type inference, and we can 
thus focus on measuring the relevance of different visual 
cues for answering various types of questions.

As baseline features, we consider the generic fc7 features from a VGG-16 trained for object classification on ImageNet and extracted from 
the whole image. To improve upon this representation, we learn other classification models on 
specialized datasets and then use them to extract ``domain expert'' features from different image regions as well as 
from the whole images. 
More specifically, we employ a scene prediction network trained on the MIT Places dataset~\citep{zhou2014learning}, 
person action networks trained on the Human Pose MPII~\citep{MPII} and Humans Interacting with Common Objects 
(HICO)~\citep{HICO} datasets, a person attribute network and an object color network trained on the Flickr30K 
Entities dataset~\citep{Flickr30kEntities_JOURNAL}. 

Together with the question types, Figure~\ref{fig:table_overview} also shows which combination of cues is used 
in each case. Note that the need to attend specific image regions is because certain question types provide ground truth bounding boxes of interest 
with the question, or because for other questions without provided boxes, the putative answers mention persons and objects. 
As an example, consider the Interestingness question in Figure \ref{fig:table_overview}  (question type 3). 
Two of the candidate answers for the most interesting aspect of this image are \emph{the girl} and \emph{firefighters}. 
In order to score these answers, we need to determine whether they 
actually exist in the image and localize the corresponding entities, if possible. To this end, we utilize an automatic bounding box selection scheme which starts with 
candidate boxes produced by state of the art person and object detectors~\citep{liu15ssd,ren2015faster} and scores
them using a region-phrase model trained on the Flickr30K Entities dataset~\citep{Flickr30kEntities_JOURNAL}. 
The highest-scoring region for a phrase contained in an answer provides spatial support for feature extraction, 
and the region-phrase scores are also used as a component of the overall answer score.
On the other hand, if persons or objects appear in the image but they are neither localized by the question 
nor named in any of the answers (see question type 1 and 2) we simply consider the image as a whole.

Each classification model used by us for feature extraction is able to predict a large vocabulary of 
semantically meaningful terms from an image: close to 200 scene categories, 1000 actions and person-object interactions, 
300 person attribute terms, and 11 colors. 
Figure~\ref{fig:overview} shows four question types from Figure~\ref{fig:table_overview} and the answer predicted by our system, as well as the intermediate predictions of our scene, action, attribute, and color 
feature networks. The outputs of these networks are semantically interpretable and can help to understand why our system 
succeeds or fails on particular questions. 
We can observe that in the Scene question example of Figure~\ref{fig:overview}, the top scene label predictions from our 
Places network (\emph{train-station-platform}, \emph{train-railway}, \emph{railroad-track}) are very similar to the correct answer 
(\emph{train station}). For the Person's Activity question, our action network cannot predict the correct activity 
(\emph{carrying a chair}) even though it corresponds to an existing class; nevertheless, it is able to predict a 
sufficiently close class (\emph{carry-suitcases}) and enable our image-text embedding method to select the correct answer.

To compute the compatibility between each of our network outputs and a candidate answer sentence or phrase, we train
a normalized Canonical Correlation Analysis (nCCA)~\citep{gong2014multi} model which maps the visual and textual 
features to a joint embedding space, such that matching input pairs are mapped close together.
More specifically, we train one nCCA model per cue, and in order to linearly combine scores from different nCCA models 
we solve an optimization problem that learns the best set of cue-specific weights.

Our high-level approach is described in Section~\ref{sec:approach}.
All the information about the used cues are provided in Section~\ref{sec:specific_cnn_models}, while the automatic bounding box selection scheme for localized feature extraction
is explained in Section~\ref{subsec:image_region_selection}. The details of our
score combination scheme is in Section~\ref{subsec:multi_cue_integration}. 
Section~\ref{sec:experiments} presents our experimental results, which show that using multiple features 
helps to improve accuracy on all the considered question types. 
Our results are state of the art, outperforming the original Madlibs baseline~\citep{VisualMadlibs}, 
as well as a concurrent method~\citep{ashkan16bmvc}.

A preliminary version of this work has appeared in BMVC~\citep{tommasi16bmvc}. The journal version includes 
(1) a more detailed description of the different cues used for each question type, 
(2) a statistical analysis of the coverage our cues provide for different types of Visual Madlibs questions (Section~\ref{sec:statistics})
(3) a principled scheme to learn an optimal weighted combination of multiple features,
(4) extensive qualitative examples to better illustrate each part of the proposed approach,
(5) a study on learning across tasks: we investigate the effect of training embedding models over multiple joint question 
types (Sections~\ref{sec:cross}) and of training the model on one question type but testing it on a different 
one (Sections~\ref{sec:shared}).

The Visual Madlibs dataset project webpage has been updated with the validation set created for our experiments: 
\url{http://tamaraberg.com/visualmadlibs/}. The deep network models used to predict various features are available at 
\url{http://vision.cs.illinois.edu/go/madlibs_models.html}.

%\clearpage
\section{Related Work}
\label{sec:related}

\textbf{Visual Question Answering.}
In the task of Visual Question Answering (VQA), natural-language questions about an image are posed to a system, and the 
system is expected to reply with a short text answer. This task extends standard detection, classification, and image captioning, 
requiring techniques for multi-modal and knowledge-based reasoning for visual understanding.
Initially proposed as a ``Visual Turing Test'' \citep{VisualTuringTest}, the VQA format has been enthusiastically 
embraced as the basis for a number of tailored datasets and benchmarks. The DAQUAR dataset \citep{malinowski14nips} 
is restricted to indoor scenes, while a number of more general datasets are based on MSCOCO images \citep{MSCOCO}, 
including COCO-QA~\citep{COCOQA}, Baidu-FM-IQA \citep{baiduVQA}, VQA \citep{VQA}, Visual7W \citep{zhu2016cvpr} and 
Visual Madlibs \citep{VisualMadlibs}. Question-answer pairs can be generated automatically by NLP tools \citep{COCOQA}, 
or created by human workers \citep{baiduVQA,VQA,zhu2016cvpr,VisualMadlibs}. 

Assessing the quality of automatically generated free-form answers is not straightforward and in most of the
cases, it reduces to evaluating the predicted probability distribution on a fixed output space made by the 1000 most
common answers of the used dataset ~\citep{emnlp16,andreas16cvpr,attentionQA,dualnet,CVPR2017WangVQA}.
Alternatively, several VQA benchmarks are provided with a multiple-choice setting where performance can 
be easily measured as the percentage of correctly answered questions.

Among automatic methods for VQA, many combine CNNs and Long Short-Term Memory (LSTM) networks to encode the questions and output the answer ~\citep{baiduVQA,malinowski2015ask,DeepCompositionalQA}. 
Recent approaches also emphasize the need for attention mechanisms for text-guided analysis of images. Such attention mechanisms can be learned, or hard-coded. Attention can be learned by using networks that predict which regions of the image are useful~\citep{xuARXIV2015,attentionQA,shihCVPR2016} and then extracting features from those regions. Hard-coded mechanisms take as input the image regions that need to be attended~\citep{zhu2016cvpr,IlievskiYF16}. Some works also use co-attention models that
exploit image regions together with word, phrase, and sentences \citep{CVPR2017WangVQA} or 
high-level concepts \citep{Yu_2017_CVPR}. In contrast to these works, our method first ranks which regions of the image are useful to the question at hand using a retrieval model, and then passes on features extracted from the useful regions to the nCCA embedding models, which select the most correct answer.

\vspace{2mm}
\noindent
\textbf{Fill-In-The-Blank Questions.}
Instead of asking explicit questions, \eg, starting with who, what, where, when, why, 
which \citep{zhu2016cvpr}, we can ask systems to fill in incomplete phrases within declarative sentences. This is the strategy behind Visual Madlibs. As stated in the Introduction and shown in Figure \ref{fig:table_overview}, Visual Madlibs questions come in twelve distinct types, some with provided regions of interest. The fact that each question has a well-defined type and structure that is known {\em a priori} makes the Visual Madlibs a more controlled
task than general VQA, enabling us to reason up front about the types of features and processing needed to answer a given question. At the same time, due to the broad coverage and diversity of these question types, we can expect 
the cues that are useful for solving Visual Madlibs to also be useful for general VQA. 

Visual Madlibs consists of 360,001 targeted natural language descriptions for 10,738 MSCOCO images, and fill-in-the-blank multiple choice questions are automatically derived from these descriptions. For each description type, the number of questions ranges between 4,600 and 7,500 
and the descriptions contain more than 3 words on average.
This makes Visual Madlibs notably different from VQA~\citep{VQA} and COCO-QA~\citep{COCOQA} datasets, which still have a multi-choice answer setting but the
majority of the answers contain a single word (see \cite{zhu2016cvpr}, Table 1). An additional unique characteristic of Visual Madlibs is in the choice of the distractor (incorrect) answers, which have two levels of difficulty: Easy and Hard. In the Easy case, the distractors are chosen randomly, while for the Hard case, they are selected from the descriptions of images containing the same objects as the test image, with similar number of words as the correct answer, but not sharing with it any non-stop words.

Existing methods for answering Madlibs questions \citep{ArunARXIV,ashkan16bmvc,VisualMadlibs} have mainly used Canonical Correlation Analysis (CCA)~\citep{Hardoon04,Hotelling36} and normalized CCA (nCCA)~\citep{gong2014multi} to create a multi-modal embedding where the compatibility of each putative answer with the image is evaluated. \cite{ashkan16bmvc} have proposed CNN+LSTM models trained on Visual Madlibs, but these were not as accurate as CCA.
The same authors have also shown that the fill-in-the-blank task benefits from a rich image representation obtained 
by detecting several overlapping image regions, potentially containing different objects, and then average-pooling 
the CNN features extracted from them.  
This representation is able to cover the abundance of image details better 
than standard whole-image features, but it uses the same kind of descriptor at all image locations. In Section \ref{sec:multi_cue_exp}, we will demonstrate that our approach of using multiple specialized descriptors outperforms \citep{ashkan16bmvc}.% for most question types.

\vspace{2mm}\noindent
\textbf{Integrating External Knowledge Sources.} 
Understanding images and answering visual questions often requires heterogeneous 
prior information that can range from common-sense to encyclopedic knowledge. To cover this need, some works integrate different knowledge sources either by leveraging training data with a rich set of different labels, or by exploiting textual or semantic resources such as DBpedia \citep{dbpedia}, ConceptNet \citep{conceptnet} and WebChild \citep{webchild}.

The approach adopted by \cite{KB_feifei_ARXIV} learns a Markov Random Field model on scene categories, attributes, and affordance labels over images from the SUN database \citep{xiao2010sun}. While this approach is quite powerful on the image side, the lack of natural language integration limits the set of possible questions that may be asked. 

The method of \cite{wu2016image} starts from multiple labels predicted from images 
and uses them to query DBpedia. The obtained textual paragraphs are then coded as a feature through Doc2Vec \citep{Doc2Vec} and used to generate answers through an LSTM. A more sophisticated technique is proposed by \cite{FVQA} for an image question task that involves only answers about common-sense knowledge: the information extracted from images and knowledge-based resources is stored as a graph of inter-linked RDF triples \citep{RDF} and an LSTM is used to map the free-form text questions to queries that can be used to search the knowledge base. The answer is then provided directly as the result of this search, avoiding any limitations on the vocabulary that would otherwise be constrained by the words in the training set. Though quite interesting, both these approaches still rely on ImageNet-trained features, missing the variety of visual cues that can be obtained from networks tuned on tasks other than object classification.

As explained in the Introduction, our own approach to integrating external knowledge relies on training ``expert'' networks on specialized datasets for scenes, actions and attributes. As one of the components of our approach, 
we use the CNN action models developed in our ECCV 2016 paper \citep{ArunARXIV}, where we applied these models 
to Person Activity and Person-Object Relationship questions (types 7 and 9) only.

\section{Overview of the Approach}
\label{sec:approach}
To tackle multiple-choice fill-in-the-blank question answering, we need a model that is able to evaluate the compatibility of each available answer choice ($a_1,\ldots,a_N$) with the image and question pair ($I,q$). This necessitates a cross-modal similarity function that can produce a score $s(I,q,a)$ taking into consideration global (whole image to whole answer) and local (image region to phrase) correspondences, as well as multiple visual cues. Our model has three main components: the image representation, the text representation, and a formulation for the cross-modal joint space and scoring function. 
\smallskip

\noindent{\bf Representing the images.} We introduce several feature types that depend on the question $q$ and possibly on the specific answer choice $a$. This dependence is made explicit by choosing how to localize the feature extraction (\emph{where} to compute the features) and \emph{which features} to extract. Broadly speaking, we have the following four types of features, each represented by networks described detail in Section \ref{sec:specific_cnn_models}.\smallskip

\begin{itemize}
\item {\bf Global image cues:}
For all question types, we extract features from the whole image using our VGG ImageNet and Places networks (see Section \ref{sec:specific_cnn_models} for details).\smallskip

\item {\bf Cues from automatically selected boxes:} 
Question types 3-5 (Interestingness, Past, and Future) do not come with any ground truth person or object boxes, but people and objects are often mentioned in candidate answers (see examples in Figure \ref{fig:table_overview} and statistics in Section \ref{sec:statistics}). We parse the candidate answers for mentioned entities and attempt to localize them using the procedure described in Section \ref{subsec:image_region_selection}. Having found the best matching image region(s) for each mentioned entity, we extract specific features depending on the nature of the entity. In particular, for people, we extract bounding box ImageNet features as well as action and attribute features, and for objects, we extract bounding box ImageNet features only.\smallskip

\item {\bf Cues from provided person boxes:}
When dealing with person-centric questions (Types 6-9), we extract features from the person bounding box provided 
with the question. These include generic ImageNet features as well as features from our action and attribute networks
\smallskip

\item {\bf Cues from provided object boxes:}
For object-centric questions (Types 9-12), we extract features from the object bounding box provided with the question using our ImageNet and color networks.%\footnoteref{note1}.
\end{itemize}

%\red{
As is clear from the above, question types 6-12, by construction of the Madlibs dataset, come with target object and person bounding boxes. %For questions that do not provide such annotations, we detect object and person bounding boxes and extract features from them.
For these question types, we did not compare performance of automatically detected vs. provided ground truth bounding boxes. Such an
experiment was performed in \citep{VisualMadlibs} 
using boxes detected by RCNN and did not show any significant difference in the performance for multiple-choice question answering. Their result indicates that detectors such as RCNN or improved methods~\citep{ren2015faster,liu15ssd} give good enough object localizations for the purposes of our end task.
A small change in the region from which features are extracted does not have a significant impact on the final question answering accuracy. On the other hand, question types 3-5 represent a more challenging case in that no target bounding boxes are provided and we will address this case at length in Section \ref{subsec:image_region_selection}.
%}
\smallskip

\noindent{\bf Representing the answers.}
Compared to our visual representation, our text representation is quite elementary.
We employ the 300-dimensional word2vec embedding trained on the Google News dataset~\citep{Word2Vec}. 
Candidate answers are represented as the average of word2vec vectors over all the words. We represent out-of-vocabulary words using the null vector, and do not encode question prompts as they are identical for all questions of the same type (e.g., ``the place is...'').
Even in the cases where the prompt contain image-specific words (\ie objects in Person-Object Relationship 
and Object's Affordance questions), adding them to the answers' representation do not introduce discriminative information, on the contrary, preliminary experiments indicated that they contribute to make the answers more similar to each other reducing the correct answer selection performance.

\smallskip

\noindent{\bf Cross-modal embedding and scoring function.}
To learn a mapping from image and text features into a joint embedding space, we adopt normalized Canonical Correlation Analysis (nCCA) \citep{gong2014multi}. For each question type, we obtain one or more nCCA scores for one or more cues corresponding to that type, and then form the final score as a linear combination of the individual scores with learned weights. Our cue combination and weight learning approaches
are described in Section \ref{subsec:multi_cue_integration}. Note that in the rest of the paper, any references to CCA models refers to nCCA models, unless otherwise specified.

\section{Cue-Specific Models}
\label{sec:specific_cnn_models}
%!TEX root = main.tex
\begin{figure*}[tb]
	\includegraphics[width=\linewidth]{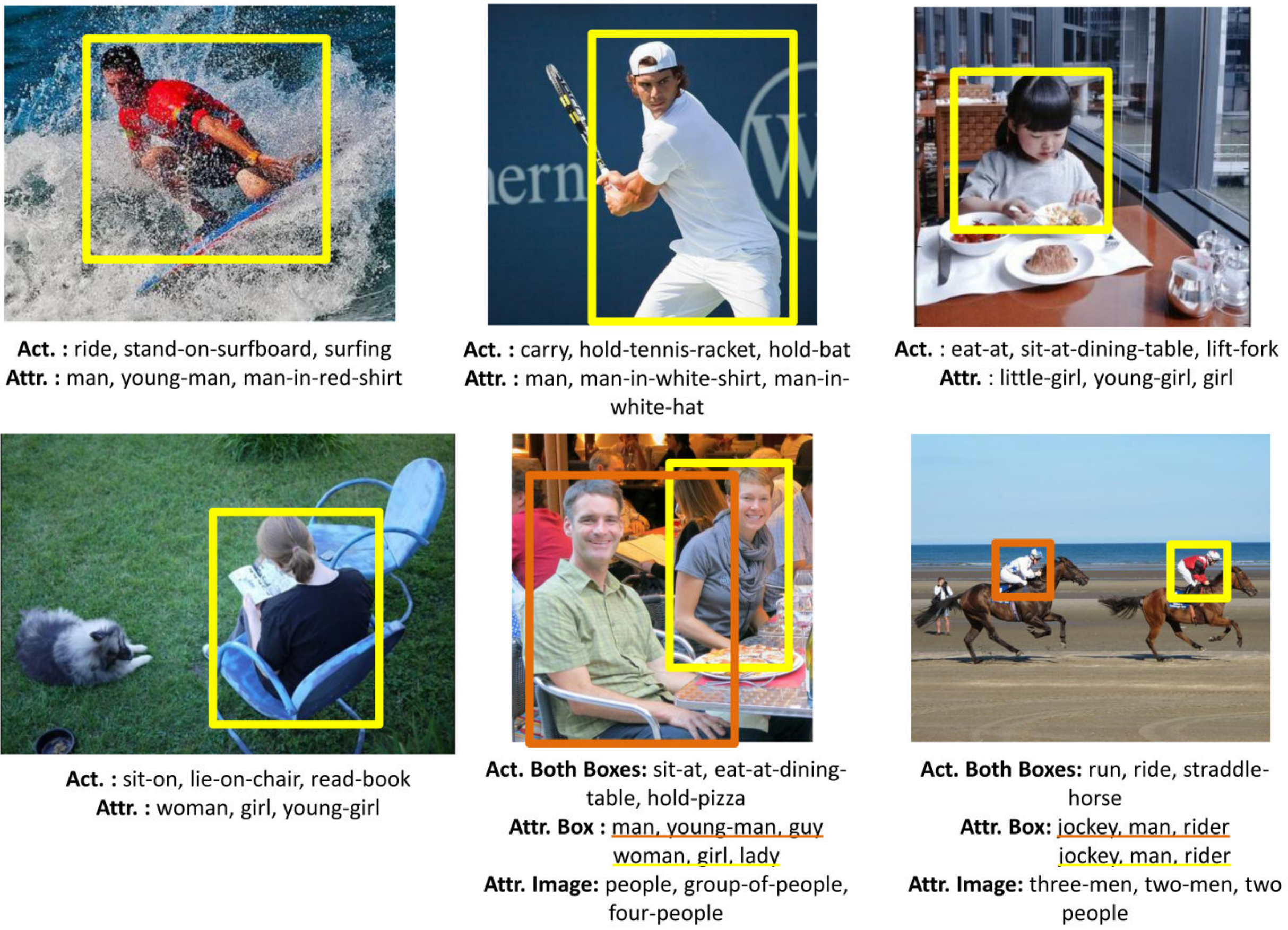}\vspace{3mm}
	\includegraphics[width=\linewidth]{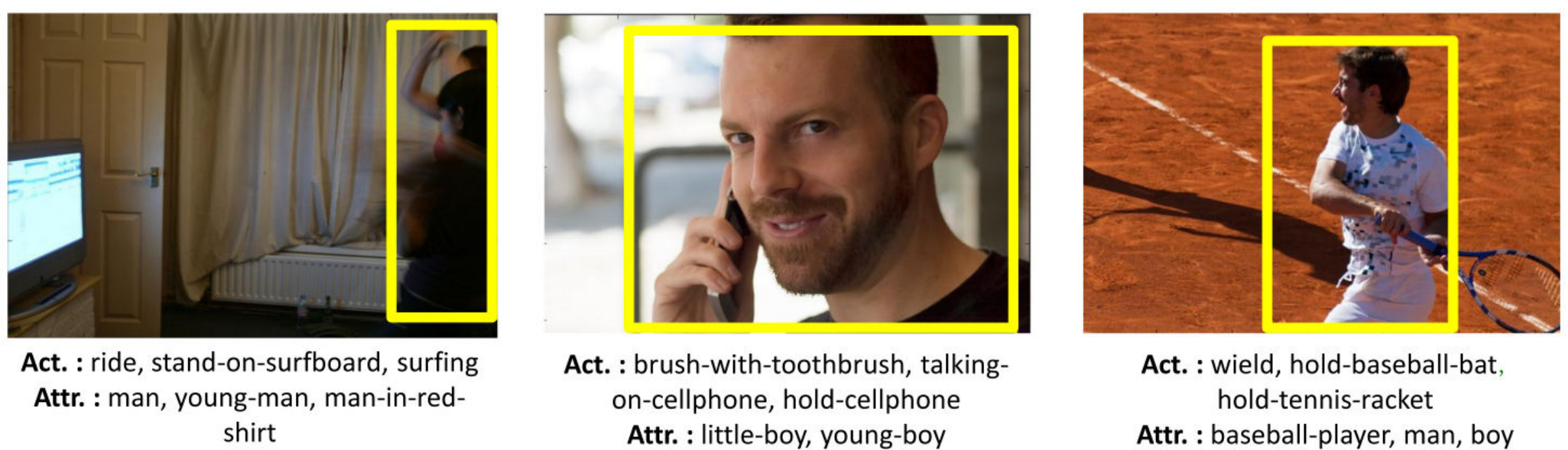}
	\caption{Top three predicted person actions (Act.) and attributes (Attr.) for a few sample images. In the case of multiple people in an image, 
    we specify the actions and attributes for specific boxes (underlined with the color of the box) as well as attributes for the whole image (Attr. Image). 
    In the last row of images we show cases where action and attribute recognition fails.} 
    %({\bf Both} means that both of the people in the image have the same action/attribute predictions.)}
	\label{figure:act_attr_preds}
\end{figure*}

This section provides details of our cue-specific networks. For a complete summary of which networks are used for which question types, refer back to Figure \ref{fig:table_overview}.\smallskip

\noindent
{\bf ImageNet network.} For all question types, we use the output of the VGG-16 network~\citep{simonyanVGG} trained on 1000 ImageNet categories as our baseline 
global feature. We obtain a 4096-dimensional feature vector by averaging fc7 
activations over 10 crops from the whole image. The same network is also used to 
extract features from image regions: in this case we indicate that it is 
a local cue, by specifying in the following tables and figures that it originates from a Person or Object bounding
box.\smallskip

\noindent
{\bf Places network.} We also use a global scene feature for each question type, derived from the Places VGG-16 network~\citep{zhou2014learning}. The MIT Places dataset contains about 2.5 million images belonging to 205 different scene categories. As with the baseline network, the Places network gives us 4096-dimensional fc7 features averaged over 10 crops.\smallskip

\noindent
{\bf HICO/MPII Person action networks.} To represent person boxes for question types 3-9, we start by passing the boxes resized to  $224 \times 224$ px as input to the generic ImageNet network. In order to obtain a more specialized and informative representation, we also use action prediction networks trained on two of the largest currently available human action image datasets:  HICO~\citep{HICO} and the MPII~\citep{MPII}. HICO has 600 labels for different human-object interactions, \eg ride-bicycle or repair-bicycle; the objects involved in the actions belong to the 80 annotated categories of the MSCOCO dataset~\citep{MSCOCO}. The MPII dataset has 393 categories, which include interactions with objects as well as solo human activities such as walking and running. 

We employ the CNN architecture introduced in our previous work~\citep{ArunARXIV}, which currently holds state of the art classification accuracy on both the action datasets. This architecture is based on VGG-16 and it fuses information from a person bounding box and from the whole image. At training time it uses multiple instance learning to account for lack of per-person labels on the HICO dataset and a weighted loss to deal with unbalanced class distributions on both HICO and MPII. The model uses a weighted logistic loss in which mistakes on positive examples are weighted ten times more than the mistakes on negative examples, in order to offset the lack of balance in the dataset.

At test time, the network of \cite{ArunARXIV} needs a person bounding box to provide a region of interest for feature extraction. For question types 6-9, these boxes are given in the ground truth. For question types 3-5, no boxes are given, so we use the automatic bounding box selection procedure that will be described in Section \ref{subsec:image_region_selection}. In case of multiple people in an image, we run the network independently on each person and then average-pool the features. In case no person boxes are detected, we use the whole image as the region of interest. 

Figure \ref{figure:act_attr_preds} presents some examples of class predictions of the action networks. For various versions of our cue combination strategies, as described in Section \ref{subsec:multi_cue_integration}, we will use either the fc7 activations of this network or the class prediction logits (inputs to the final sigmoid/softmax layer). \smallskip

\noindent
{\bf Person attribute network.} For question types 3-9, alongside generic ImageNet features and activity features described above, we also extract high-level features based on a rich vocabulary of describable person attributes. To create such a vocabulary, we mine the Flickr30K Entities dataset~\citep{Flickr30kEntities_JOURNAL} for noun phrases that refer to people and occur at least 50 times in the training est. This results in 302 phrases that cover references to gender (man, woman), age (baby, elderly man), clothing (man in blue shirt, woman in black dress), appearance (Asian man, brunette woman), multiple people (two men, group of people), and more. An important advantage of our person attribute vocabulary is that it is an order of magnitude larger than those of other existing datasets~\citep{Sudowe2015Parse27k, bourdev2011BAPD}. On the down side, attributes referring to males (\eg man, boy, guy, \etc) occur twice as often as those referring to females (\eg woman, girl, lady, \etc), and the overall class distribution is highly unbalanced (i.e., there are a few labels with many examples and many classes with just a few examples each). 

We train a Fast-RCNN VGG-16 network~\citep{girshick2015fast} to predict our 302 attribute labels based on person bounding boxes (in case of group attributes, the ground truth boxes contain multiple people). To compensate for unbalanced training samples, just as for the action networks, we use a weighted logistic loss that penalizes mistakes on positive examples ten times more than on negative examples. Unlike our action prediction network, our attribute network does not use global image context (we found that attribute predictions are much more highly localized and tend to be confused by outside context) and it predicts group attributes given a box with multiple people (such boxes naturally exist in the Flickr30K Entities annotations). As our labels are derived from natural language phrases, we manually grouped and ignored predictions on labels which could be simultaneously true but are not annotated in the dataset. For example, if a bounding box is referred to as \emph{he}, \emph{man in blue shirt}, \emph{older man}, or \emph{bald man}, related labels such as \{\emph{man}, \emph{gentleman}, \emph{guy}, \emph{man in hard hat}, \emph{asian man}\} might also be true. Essentially, the presence of a label such as \emph{he} does not conclusively indicate the absence of all other labels, such as \emph{guy}, however it does indicate the absence of \emph{she}, or \emph{woman}. We manually created four such label groups representing man, woman, boy, and girl. If a label belongs to a given group, labels from all other groups can be safely considered as negatives, while labels within a group can be ignored while computing the training loss.

\begin{figure*}[tb]
    \includegraphics[width=\linewidth]{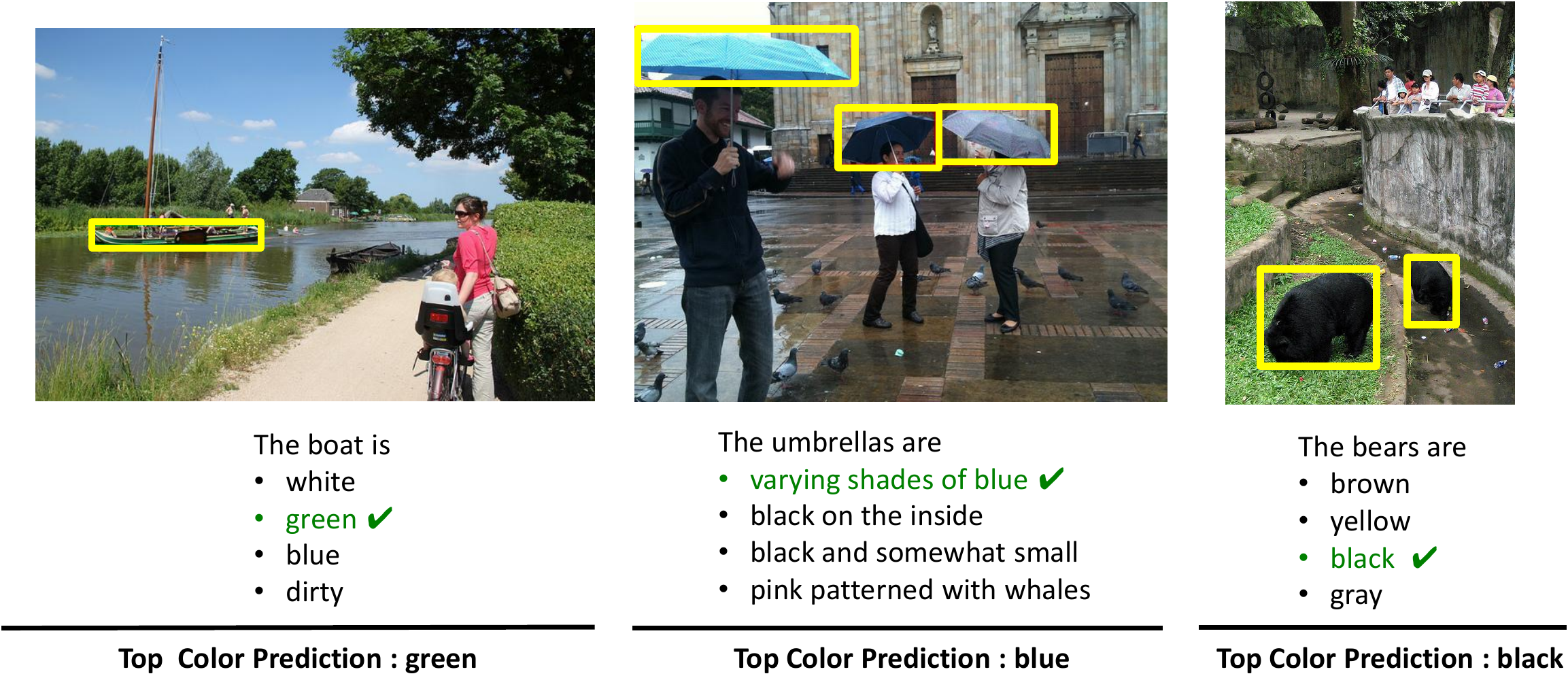}\vspace{2mm}
    \includegraphics[width=\linewidth]{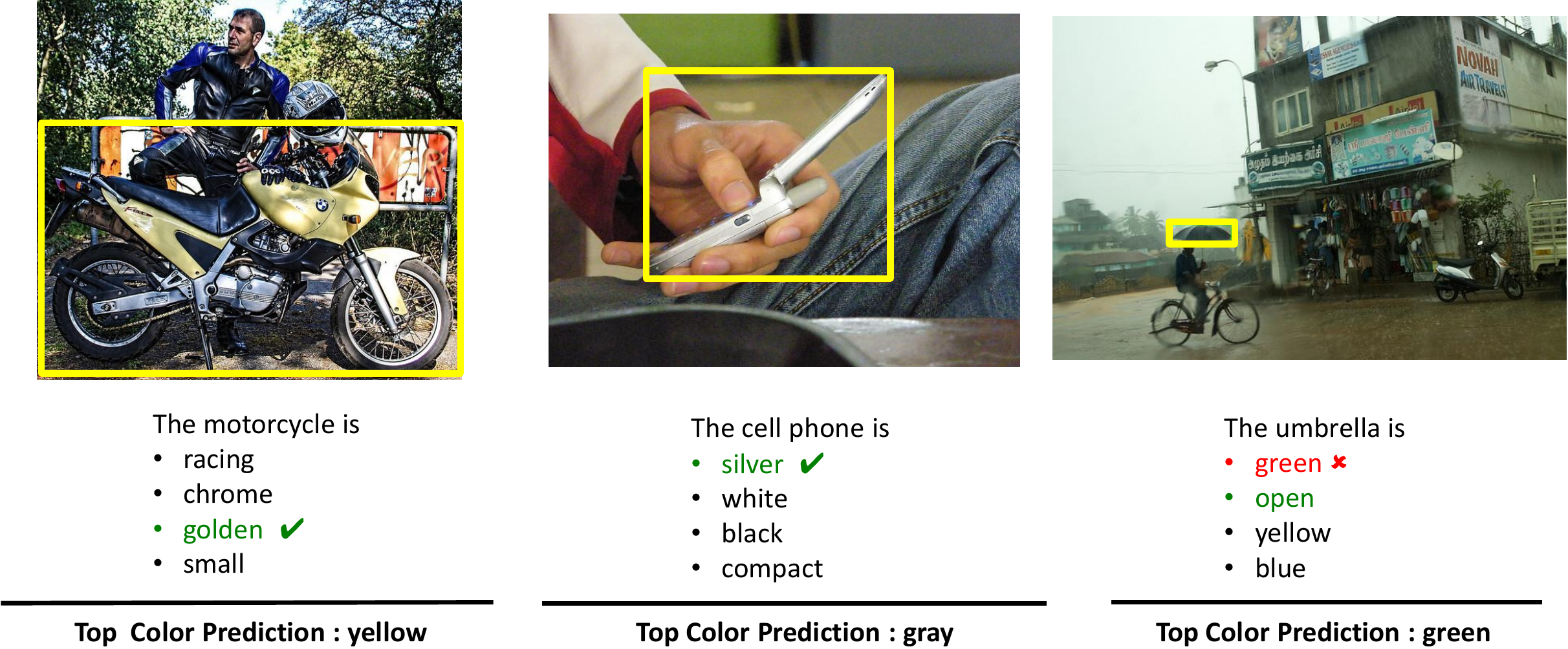}
 	\caption{
    Examples of Object's Attribute questions with the top prediction of our Color network underneath. Even if the color mentioned in the answer is not among the ones predicted by our model, it can still be relevant (second row, first two images). The bottom right image is a failure case where the predicted color leads to the wrong answer.}
	\label{figure:color}%\vspace{-0.5mm}
\end{figure*}

To give a quantitative idea of the accuracy of our person attribute prediction, the mAP of our network on the phrases of the Flickr30k test set that occur at least 50 (resp. 10) times is 21.98\% (resp. 17.04\%). We observe the following APs for some frequent phrases: man - 53.8\%, woman - 51.3\%, couple - 35.4\%, crowd - 36.1\%. It should be noted that these numbers likely underestimate the accuracy of our model. For one, they are based on exact matches and do not take synonyms into account. Moreover, there is a significant sparsity problem in the annotations, as numerous attribute phrases may be applicable to any person box but only a few are mentioned in captions. Qualitatively, the attribute labels output by our network are typically very appropriate, as can be seen from example predictions in Figure~\ref{figure:act_attr_preds}. 

At test time, to obtain person bounding boxes from which to extract attribute features, we follow the same procedure as for the action networks described above. In case of multiple people boxes, the outputs of the attribute network are average-pooled. As with the action models, either the inputs to the final sigmoid/softmax layer or the fc7 activations can be used for the downstream question answering task (refer to Sections \ref{subsec:multi_cue_integration} and \ref{sec:experiments} for details). \smallskip

\noindent{\bf Color network.} 
As described in Section \ref{sec:approach}, we extract object-specific cues for automatically detected boxes on question types 3-5 (Interestingness, Past, Future), as well as for provided focus boxes for question types 9-12. For all of those object boxes, just as for person boxes in question types 3-9, we extract generic ImageNet features from the bounding boxes. To complement these, we would also like to have a representation of object attributes analogous to our representation of person attributes. However, it is much harder to obtain training examples for a large vocabulary of predictable attributes for non-human entities. Therefore, we restrict ourselves to color, which is visually salient and frequently mentioned in Visual Madlibs descriptions, and is not captured well by networks trained for category-level recognition~\citep{Flickr30kEntities_JOURNAL}. We follow~\cite{Flickr30kEntities_JOURNAL} and fine-tune a Fast-RCNN VGG-16 network to predict one of 11 colors that occur at least 1,000 times in the Flickr30K Entities training set: black, red, blue, white, green, yellow, brown, orange, pink, gray, purple. This network is trained with a one-vs-all softmax loss. The training is performed on non-person phrases to prevent confusion with color terms that refer to race. For our color feature representation we use the 4096-dimensional fc7 activation values extracted from the object bounding box.

Quantitative evaluation of a color network similar to ours can be found in~\cite{Flickr30kEntities_JOURNAL}. The examples in Figure \ref{figure:color} provide a qualitative illustration of the color network outputs and indicate how color predictions may be helpful for answering Object's Attribute Visual Madlibs questions.

Note that we extract color features only from provided object boxes for questions 9-12. For questions 3-5, color is mentioned far more rarely in candidate answers; furthermore, automatically detected object boxes are much more noisy than person boxes making the color cues correspondingly unreliable.

\section{Image Region Selection}
\label{subsec:image_region_selection}
Madlibs questions on Interestingness, Past, and Future do not provide a target image region. 
Consider the Future example in Figure~\ref{fig:overview}, where each of the four candidate answers mentions a person and an object: \emph{she put down the cat}, \emph{the bride dropped the bouquet}, and so on. In order to pick the right choice, we need to select the best supporting regions for each of the entity mentions (\emph{she}, \emph{cat}, \emph{bride}, \emph{bouquet}) and use the respective matching scores as well as the features extracted from the selected regions as part of our overall image-to-answer scoring function. 

We first parse all answers with the Stanford parser \citep{SocherEtAl2013:CVG} and use pre-defined vocabularies to identify noun phrase (NP) chunks referring to a person or to an object. Then we apply the following region selection mechanisms for mentioned people and objects, respectively.\smallskip

\noindent{\bf Person Box.} We first detect people in an image using the Faster-RCNN detector~\citep{ren2015faster} with the default confidence threshold of 0.8. We discard all detected boxes with height or width less than 50 pixels since we find experimentally that these mainly contain noise and fragments. We also consider the smallest box containing all detected people, to account for cues originating from multiple people. Given the image and an answer, we attempt to select the box that best corresponds to the person mention in the answer. To this end, we train a {\bf Person CCA model} on the val+test set of Flickr30k Entities using person phrases (represented by average of word2vec) and person box features (302-dimensional vectors of predictions from our person attribute network of Section \ref{sec:specific_cnn_models}). As a lot of answer choices in the Madlibs dataset refer to people by pronouns or collective nouns such as \emph{he}, \emph{she}, \emph{they}, \emph{couple}, we augmented the training set by replacing person phrases as appropriate. For example, for phrases such as \{\emph{man}, \emph{boy}, \emph{guy}, \emph{male}, \emph{young boy}, \emph{young man}, \emph{little boy}\}, we added training samples in which these phrases are replaced by \emph{he} (and the same for \emph{she}). Similarly, additional examples were created by replacing \{\emph{people}, \emph{crowd}, \emph{crowd of people}, \emph{group of people}, \emph{group of men}, \emph{group of women}, \emph{group of children}\}, \etc, with \emph{they}, and \{\emph{two men}, \emph{two women}, \emph{two people}\} with \emph{couple}.

\begin{figure*}[tb]
	\includegraphics[width=\linewidth]{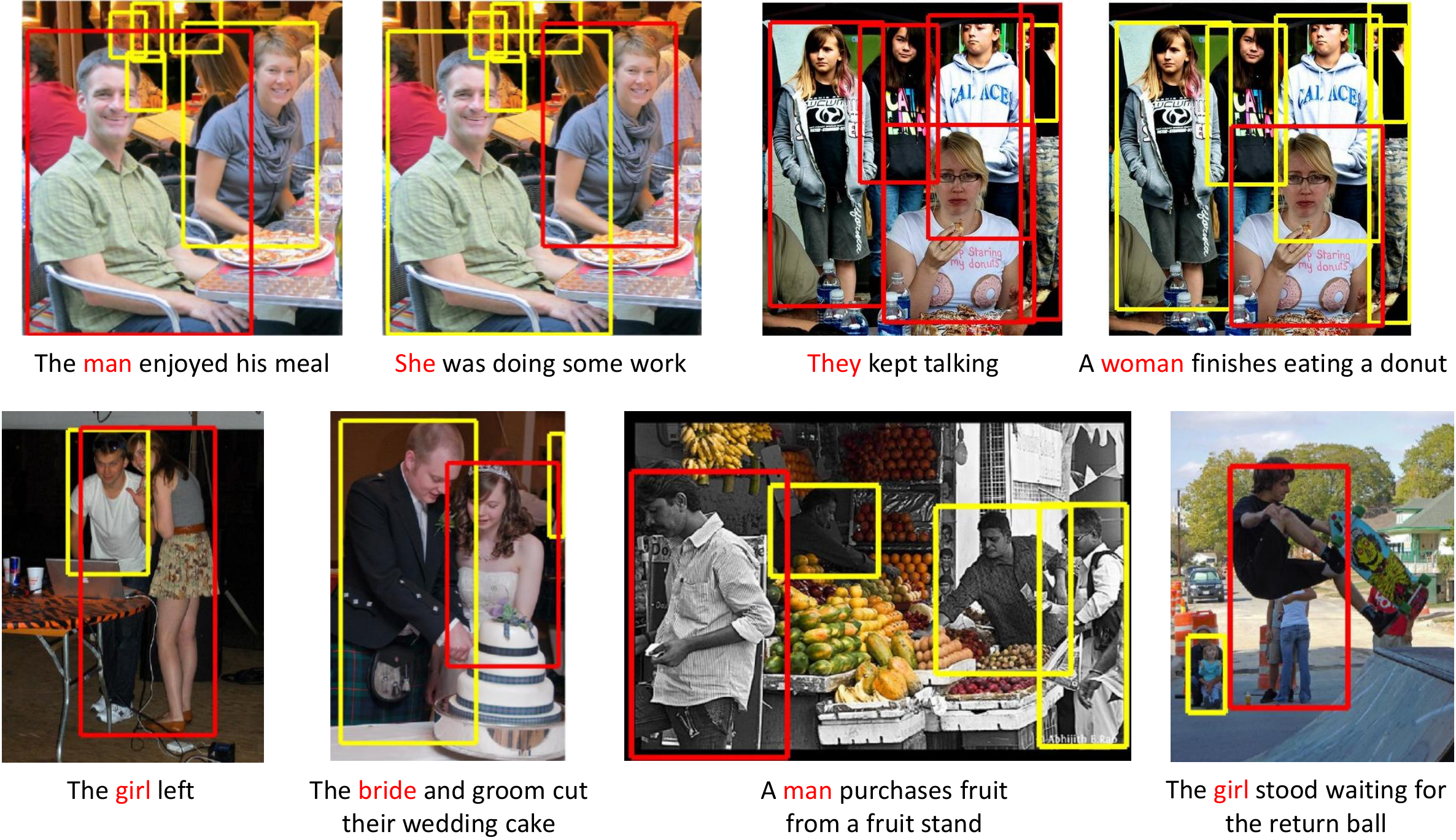}
	\caption{Examples of selected person boxes based on person phrases. The person phrases are highlighted in red font and the corresponding selected boxes are also colored red. The yellow boxes are discarded either because they do not match the person mentioned in the phrase or because they are below the size threshold. In the third example from the left in the top row, CCA selects the overall box, thus all the person-specific boxes are colored red with the exception of the top right one which is discarded as it is below the size threshold. The last two images in the second row are failure cases.}
	\label{figure:box_selection}
\end{figure*}

Given the trained Person CCA model, we compute the score for each person phrase from the candidate answer and each candidate person box from the image, and select the single highest-scoring box. A few example selections are shown in Figure~\ref{figure:box_selection}. In case no words referring to people are found in a choice, all person boxes are 
selected.\footnote{Note that the images of the Visual Madlibs dataset are sampled from the MSCOCO dataset \citep{MSCOCO} to contain at least one person.} The selected box provides spatial support for extracting person action and attribute cues introduced in Section \ref{sec:specific_cnn_models}; in turn, these features, together with entire candidate answers (as opposed to just the person phrases), are used to train cue-specific CCA models as will be explained in the next section. The score of the Person CCA model for the selected box will also be used in a trained combination with the cue-specific CCA scores. \smallskip

\begin{figure*}[t!]
	\includegraphics[width=\linewidth]{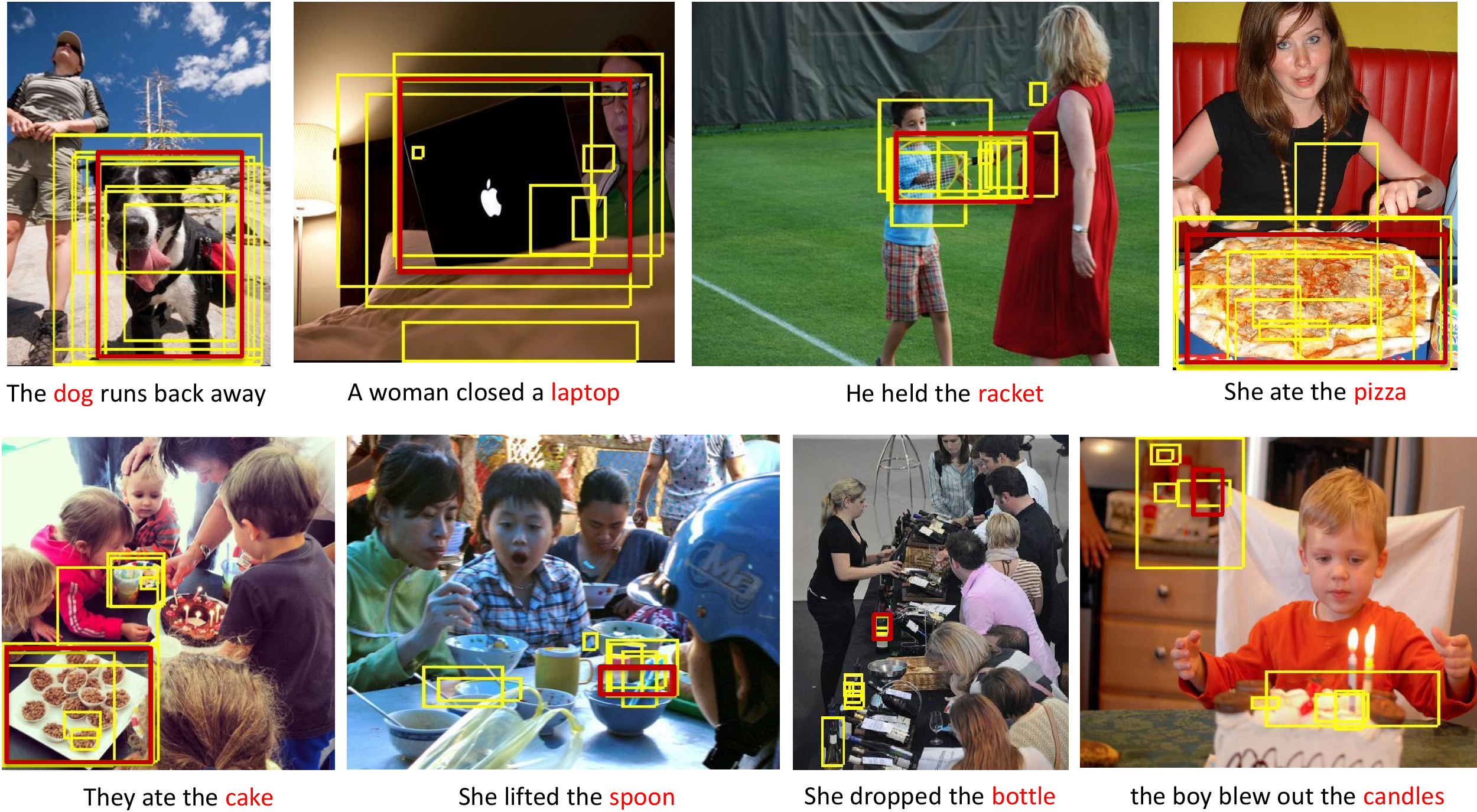} 
	\caption{Examples of selected object boxes based on object phrases (in red font). The red boxes are the top-scoring ones according to the object CCA model, while all the yellow boxes have lower scores. The top row presents correctly detected objects, while the bottom row shows failure cases.}
	\label{figure:obj_box_selection}
\end{figure*}

\noindent{\bf Object Box.}
We localize objects using the Single Shot MultiBox Detector (SSD)~\citep{liu15ssd} that has been trained on the 80 MSCOCO object categories. SSD is currently the state of the art for detection in speed and accuracy. For each Visual Madlibs image, we consider the top 200 detections as object candidates and use the {\bf Object CCA model} created for the phrase localization approach of~\citep{Flickr30kEntities_JOURNAL} to select the boxes corresponding to objects named in the sentences. This model is trained on the Flickr30k Entities dataset over Fast-RCNN fc7 features and average of word2vec features. We use the simplest model from that work, not including size or color terms. The top-scoring box from the image is used to extract object VGG features as will be explained in Section \ref{subsec:multi_cue_integration}.

Figure~\ref{figure:obj_box_selection} shows a few examples of object selection in action. As can be seen from the failure cases in the bottom row, object boxes selected by our method are less reliable than selected people boxes, since detection accuracies for general objects are much lower than for people and object boxes tend to be smaller. %An additional problem is that different answers may mention varying numbers of distinct objects. 
Therefore, instead of defining an object selection score based on the single highest-scoring region-phrase combination, as in the case of people above, we define a collective object score that will be used in the cue combination method of Section \ref{subsec:multi_cue_integration}. Inspired by a kernel for matching sets of local features~\citep{Lyu:05}, we take all of the $N=200$ object boxes from the image and the $M$ object phrases from the answer and then combine their CCA matching scores as follows: 
%\vspace{-2mm}
\begin{align} \label{eq:kernel}
K(image, answer) &= \nonumber \\
& \hspace{-2cm} \frac{1}{N}\frac{1}{M}\sum_{i=1}^N\sum_{j=1}^M \{\text{cos\_similarity}(box_i, phrase_j)\}^r~,
\end{align}
where the parameter $r$ assigns more relative weight to box-phrase pairs with higher similarity. 
We use $r=5$ in our implementation.

\section{Cue Combination}
\label{subsec:multi_cue_integration}
As described in Section \ref{sec:specific_cnn_models}, we extract several types of features from the images, aiming to capture multiple visual aspects relevant for different question types. How can we combine all these cues to obtain a single score $s(I,q,a)$ for each question, image and candidate answer?

The simplest combination technique is to concatenate 4096-dimensional fc7 features produced by each of our networks. In practice, due to the dimensionality of the resulting representation, we can only do this for a pair of networks, obtaining 8192-dimensional features. In our system, we mainly use this technique when we want to combine our baseline global ImageNet network with one other cue. 

To combine more than two features, we can stack lower-dimensional class prediction vectors (logits, or values before the final sigmoid/softmax layer). In particular, to characterize people, we concatenate the class predictions of HICO, MPII, and attribute networks, producing a compact feature vector of 1295 dimensions.

To enable even more complex cue integration, we learn CCA models on small subsets of cues and linearly combine their scores with learned weights. The following is a complete list of the individual CCA models used for our full ensemble approach:
\begin{itemize}
\item {\bf Baseline + Places:} CCA trained on concatenated fc7 features from global ImageNet- and Places-trained networks. This is used for all question types.
\item {\bf Baseline + Person Box ImageNet:} CCA trained on concatenated fc7 features from ImageNet network applied to the whole image and person box. This cue is used for question types 3-5 (on automatically selected boxes) and 6-9 (on ground truth boxes). The reason for concatenating the global and person box features is to make sure that the resulting model is at least as strong as the baseline. The same reasoning applies to the other person-specific and object-specific models below.
% \item {\bf Baseline + Person Box HICO (resp. MPII):} CCA trained on concatenated fc7 features from VGG network applied to the whole image and from the HICO (resp., MPII, attribute) network applied to the person box. Used for question types 3-9. 
% \item {\bf Baseline + Person Box Attribute:} CCA trained on concatenated fc7 features from VGG network applied to the whole image and from the person attribute network applied to the person box. Used for question types 3-9.
\item {\bf HICO + MPII + Person Attribute:} CCA trained on concatenated logit scores from HICO, MPII, and Attribute networks. Used for question types 3-9.
\item {\bf Person selection score:} Person box selection score from the Person CCA model of Section \ref{subsec:image_region_selection}. Used for question types 3-5.
\item {\bf Object selection score:} Scores from the Object CCA model of Section \ref{subsec:image_region_selection} combined using eq. (\ref{eq:kernel}). Used for question types 3-5.
\item {\bf Baseline + Object Box ImageNet:} CCA trained on concatenated fc7 features from the ImageNet network applied to the whole image and object box. Used for question types 3-5 (on automatically selected boxes) and 9-12 (on ground truth boxes).
\item {\bf Baseline + Object Box Color:} CCA trained on concatenated fc7 features from the ImageNet network applied to the whole image and color network applied to the object box. Used for question types 9-12.
\end{itemize}

To learn the combination weights, we divide the Visual Madlibs training set into an $80\%$ training subset and a $20\%$ validation subset. From the training subset, we learn the individual CCA models above using respective features and text descriptions\footnote{The Madlibs training set contains only the correct image descriptions, not the incorrect distractor choices.}. For the validation set, we create three Easy and three Hard distractors for each correct description by following the same rules originally applied to create the test set~\citep{VisualMadlibs}. 

For a particular question type, let $s^j$ indicate the CCA score obtained on the validation sample $(I,q,a)$ when using the $j$th model. We can then combine scores from all CCA models applicable to this question type as $S = \sum_j w^j s^j$. Let $S_i$ denote the combined score for each candidate answer $a_i$ for the considered sample, and $i^*$ the index of the correct choice. We define the following convex loss: 
\begin{equation}
 L(S) = \max\{1- S_{i^*} + \max_{i\neq i^*}\{S_i\}, 0\}~.
\end{equation}
This formulation assigns zero penalty when the score of the correct answer is larger by at least 1 than the 
scores of all the wrong choices. Otherwise, the loss is linearly proportional to the difference
between the score of the correct answer and the maximum among the scores of the other choices.
Over all the $k=1,\ldots,K$ validation samples, we solve
\begin{equation}%\small
 \min_{\beta} \sum_{k=1}^K L(S)_k  \quad\text{subject to}\quad \|w\|_1\leq 1~, \quad w^j\geq 0~,
 \label{eq:L1reg}
\end{equation}%\normalsize
where the constraints specify that the weights for each feature should be positive, and the 
L1-norm condition can be seen as a form of regularization which induces a sparse solution
and allows an easy interpretation of the role of each cue. Alternatively, we tried using the L2-norm and obtained slightly lower final performance. One large advantage of using the L1 norm is that the assigned weights provide good interpretability of the relevance of cues, as will be seen in Table~\ref{table:weights}.
%allowing to disregard irrelevant cues for the question at hand. 
We implemented the optimization process by using the algorithm of \cite{Duchi:2008:EPL}.

For a test question of a given type, we compute all the applicable CCA scores, combine them with the learned weights for that question type, and choose the answer with the highest combined score:
\begin{equation}
a_i^* = \argmax_{i} \{S_i\} = \argmax_{i} \left\{ \sum_j w^j s^j_i\right\}~.
\end{equation}

\section{Experiments}
\label{sec:experiments}
In Section \ref{sec:statistics}, to motivate our selection of cues for different questions, we examine the frequencies of cue-specific words in answers for each question type. 
In Section \ref{sec:single_cue_exp}, we proceed to a detailed analysis of the multiple-choice answer task when using each cue separately. Finally, in Section \ref{sec:multi_cue_exp} we evaluate the performance of the combined system.
A further analysis of our approach in cross-tasks settings is presented in Section \ref{sec:shared} and \ref{sec:cross}
where we discuss the effect of learning CCA embeddings over multiple joint question types and 
of testing the embedding on a different question type with respect to that used in training.

\subsection{Cue-Specific Category Statistics}
\begin{figure*}[tb]
\begin{center}
\resizebox{\textwidth}{!}{%
	\begin{tabular}{c@{}c@{}c@{}c@{}c}
       \includegraphics[width=\textwidth]{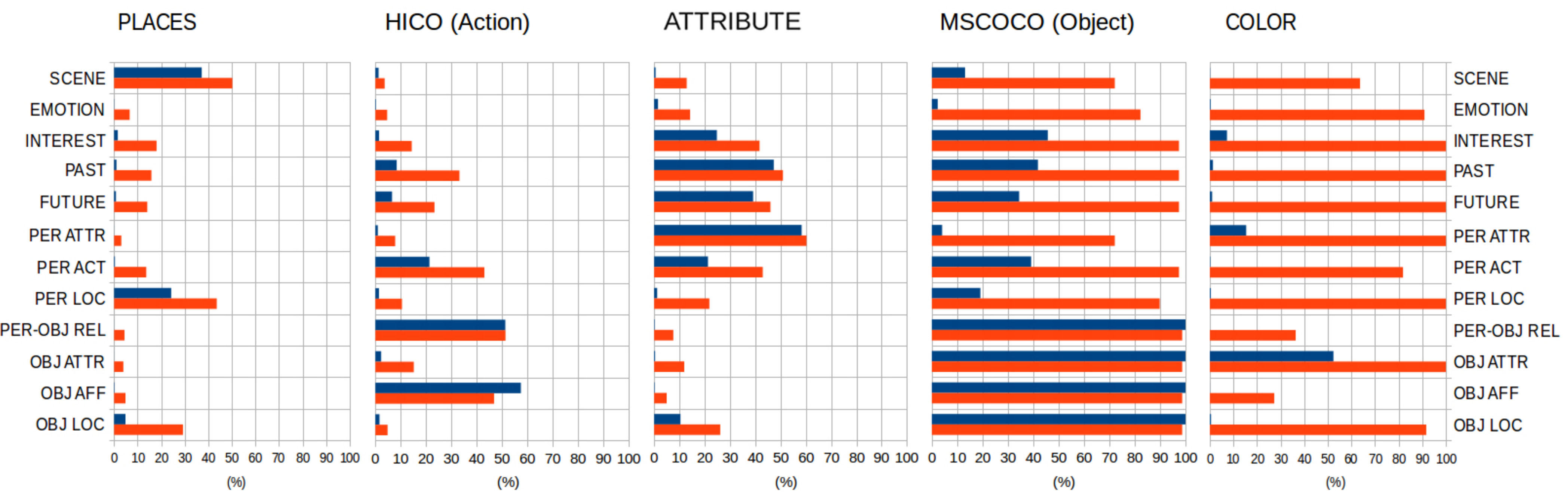}\\
    \end{tabular}
}
	\caption{{\bf{Madlibs coverage (blue)}}: indicates which percentage of the Visual Madlibs sentences 
    mentions at least one of the Places (205 classes), action (HICO, 600 classes), attribute (302 classes), MSCOCO object (80 classes) and color (11 classes) categories. {\bf{Category list coverage (red)}}: indicates which percentage of the category list is named at least once in the Visual Madlibs sentences. Both the coverage evaluations are performed by starting from the ground truth correct answers of the Visual Madlibs training set.}
	\label{figure:stat}
    \end{center}
\end{figure*}

Given the lists of 205 Places scene categories, 600 HICO action categories, 302 attribute categories, 80 MSCOCO object categories, 
and 11 color categories, we can compute the following statistics for ground truth correct answers from the training set 
(i.e., accurate descriptions) of each Visual Madlibs question type: 
\begin{itemize}
\item \textbf{Madlibs coverage} = (number of answers that mention at least one of the categories) / (total number of answers);
\item \textbf{Category list coverage} = (number of categories named at least once in the answers) / (total number of categories).
\end{itemize}
When counting the occurrences of the HICO actions, we consider past tense, continuous (-ing) and third person (-s) forms of the verbs. 
We also augment the MSCOCO object vocabulary with several word variants (e.g. bicycle, bike etc.) and singular/plural forms for 
all the objects.

Figure \ref{figure:stat} shows the resulting statistics. Not surprisingly, Places categories have the best coverage on Scene questions: 
about $37\%$ of the Visual Madlibs Scene answers mention one out of $50\%$ of the Places categories. Beyond that, about 25\% of Person's 
Location answers mention one of $40\%$ of the Places categories, and about 5\% of Object's Location answers mention one of 30\% of these 
categories. 

HICO action categories give the best coverage for Person-Object Relationship, Object's Affordance, and Person's Activity questions. 
Attribute classes play an important role for Interestingness, Past, Future, Person's Attribute, and Person's Activity questions. 
However, no more than about 50\% (resp. 60\%) of HICO Action (resp. Attribute) categories are mentioned in answers of any 
single given type. 

By contrast, more than 70\% of the MSCOCO objects appear in all the question types and 100\% of the 
Object related answers (question types 9-12) mention one of the MSCOCO categories. This is not surprising, since the 
Visual Madlibs dataset was created on top of MSCOCO images. Objects are also often mentioned by Interestingness, Past, 
Future and Person's Action answers, but are rare in all the remaining cases.

Finally, Color categories play the most important role for Object's Attribute questions: over 50\% of answers for that 
question type mention a color, and 100\% of the color names are mentioned. While a majority of the color names are also 
mentioned in all the other question types except for Person-Object Relationship and Object's Affordance, the percentage 
of answers that actually mention a color is negligible.

This analysis support a preliminary selection of the cues to use in each 
case. Since actions, attributes, objects and their colors are not named in the answers
of the Scene and Emotion question types, the visual appearance of a person/object instance in 
the images would not have any matching textual information. 
Similarly, the sparse presence of person attribute mentioned
in the Object related question (types 10-12) indicate that people are rarely pointed out
in the answers. Without a phrase that explicitly refers to an object/person instance 
we do not have a reasonable spatial support to extract local features, thus we decided
to avoid them. Finally object colors provide only a limited amount of information
due to their low coverage and to avoid further noise introduced by the object 
localization it makes sense to include them only when the object bounding box is 
provided with the question (types 9-12).

\label{sec:statistics}

\subsection{Single-Cue Results}

\begin{table*}[tb!]
	\setlength{\tabcolsep}{5pt}
	\footnotesize
	\begin{center}
    \resizebox{\textwidth}{!}{%
		\begin{tabular}{@{}c|ll||c|c|c|c|c|c|c|c@{}} 
            \hline
  {\multirow{2}{*}{Distractor}} & \multicolumn{2}{c||}{\multirow{2}{*}{Question Type}} & \multicolumn{2}{c|}{Full Image} & \multicolumn{4}{c|}{Person Box} & \multicolumn{2}{c}{Object Box}\\ 
            \cline{4-11}
            {\multirow{2}{*}{Type}}& & & Baseline & B. +  & B. + & B. + & B. + & B. + & B. + & B. +\\
            && & ImageNet      & Places      & ImageNet        & HICO  & MPII  & Attr.      & ImageNet        & Color\\ \hline 
            \multirow{12}{*}{Easy} &\multirow{5}{*}{(A)} & {1) Scene}     & 87.73 & \bf{89.04} & -- & -- & -- & -- & -- & --\\
			&                                            & {2) Emotion}	  & 48.32 & \bf{49.53} & -- & -- & -- & -- & -- & --\\                                           
			&&  {3) Interesting}       & 78.11 & 78.74 & \bf{79.59} & 79.47 & 78.55 & 79.31 & 78.86 & --\\                                            
			&&  {4) Past}		      & 79.30 & 80.34 & 80.60 & 81.54 & 80.28	& \bf{81.68} & 80.36 & --\\ 
			&&  {5) Future}            & 79.52 & 80.10 & 80.76 & \bf{82.29}	& 80.42  & 81.30 & 80.61 & --\\  
            \cline{2-11}
			&\multirow{4}{*}{(B)} & 6) Person's Attribute & 53.32 & 54.10 & 59.00& 54.10 & 55.60	& {\bf 64.50} & -- & -- \\                       
            &		                  & 7) Person's Activity    & 83.89 & 84.30 & 85.51 & {\bf 87.46} & 85.16 & 84.71 & --& --\\ 
            &		                  & 8) Person's Location  & 84.59 & {\bf 85.70} & 84.50 & 85.29 & 84.51 & 84.33 & --& --\\           
			&		                  & 9) Person-Object Relation & 71.36 & 72.07 & 72.80 & {\bf 75.77} & 73.84 & 71.10 & 74.39 & 71.55\\ 
            \cline{2-11}
			&\multirow{3}{*}{(C)} & 10) Object's Attribute  & 50.15 & 50.47	 & -- & -- & -- & -- & 57.85 & \bf{59.50}	\\ 
			&					      & 11) Object's Affordance & 80.56 & 82.76	 & -- & -- & -- & -- & \bf{87.20} & 82.32	\\ 
            &                          & 12) Object's Position	 & 67.79 & \bf{69.41} & -- & -- & -- & --	& 68.18	& 67.98	\\  
 %           \cline{2-11}
 %           &						  & Average	             & ?? & ?? & -- & -- & -- & -- & -- & --\\                                                   
            \hline   
            \multirow{12}{*}{Hard} &\multirow{5}{*}{(A)}  & {1) Scene}      & 70.94 & \bf{73.22} & -- & -- & -- & -- & -- & --\\
			&                                             &  {2) Emotion}	& 35.50 & \bf{35.77} & -- & -- & -- & -- & -- & --\\                               
			&&  {3) Interesting}       & 54.36 & 54.66 & 54.60 & 54.92 & 54.95 & \bf{56.02} & 54.33 & --\\                                            
			&&  {4) Past}		      & 53.89 & 53.95 & 55.37 & 55.09 & 54.11& \bf{55.90}& 53.78 & -- \\ 
			&&  {5) Future}            & 55.19 & 55.47 & 56.25 & 56.76 & 55.19  & \bf{57.58} & 57.02 & --\\  
            \cline{2-11}
			&\multirow{4}{*}{(B)} & 6) Person's Attribute & 42.55 & 43.11 & 48.85 & 43.06 & 45.77	& {\bf 54.64} & -- & --\\                       
            &		                  & 7) Person's Activity    & 67.56 & 68.10 & 69.47 & {\bf 71.02} & 70.03 & 68.68 & --& -- \\ 
            &		                  & 8) Person's Location  & 64.57 & {\bf 66.66} & 65.46 & 64.97 & 64.76 & 64.71 & --& --\\           
			&		                  & 9) Person-Object Relation & 54.46& 54.65 & 56.84 & {\bf 58.72} & 56.84 & 54.48 & 55.85 & 54.58 \\ 
            \cline{2-11}
			&\multirow{3}{*}{(C)} & 10) Object's Attribute          & 44.99& 45.62	 & -- & -- & -- & -- & 53.63 & \bf{54.73}	\\ 
			&					      & 11) Object's Affordance         & 64.26 & 64.50 	 & -- & -- & -- & -- & \bf{67.65} & 63.99 \\ 
            &                          & 12) Object's Position	          & 56.46 & \bf{57.56} & -- & -- & -- & --	& 57.34	& 56.43	\\  
%           \cline{2-11}            
%            &                         & Average                        & ?? & ?? & -- & -- & -- & -- & -- & --\\                                        
            \hline                     
        \end{tabular}
        }\vspace{-2mm}
	\end{center}
\caption{Accuracy on Madlibs questions with fc7 features. The Baseline ImageNet column gives performance for 4096-d fc7 outputs of the baseline network trained on ImageNet classification. For the columns labeled ``B. + X'', the baseline fc7 features are concatenated with fc7 features of different specialized networks, yielding 8192-d representations. }
	\label{table:madlibs_accuracy}
    %\vspace{-4mm}
\end{table*}

This section analyzes the performance of our individual cues listed in Section \ref{subsec:multi_cue_integration}. 
The results are presented in Table~\ref{table:madlibs_accuracy}: each question type is considered separately in the
experiments but to ease the discussion we organized the questions on the basis of their visual focus: whole image 
(types 1-5, A), person-specific (types 6-9, B) and object-specific (types 10-12, C).
The leftmost column shows the accuracy obtained with the baseline whole-image ImageNet fc7 feature. The subsequent columns show the performance obtained by concatenating this feature with the fc7 feature of each of our individual cue-specific network (as explained in Section \ref{subsec:multi_cue_integration}, the reason for always combining individual cues with the baseline is to make sure they never get worse performance).\smallskip

\noindent{\bf Whole-Image Questions.} As shown in Table~\ref{table:madlibs_accuracy}(A), using the Places features for Scene questions helps to improve performance over the ImageNet baseline. Emotion questions are rather difficult to answer but we can observe some improvement by adding Place features as well. We did not attempt to use person- or object-based features for the Scene and Emotion questions since the analysis of Section \ref{sec:statistics} indicated a negligible frequency of person- and object-related words in the respective answers.

On the other hand, for Future, Past, and Interestingness questions, people and objects play an important role, hence we attempt to detect them in images as described in Section~\ref{subsec:image_region_selection}. From the selected person boxes we extract fc7 features from four different networks: the generic ImageNet network, the HICO and MPII Action networks, and the  Attribute network trained on Flickr30K Entities. All of them give an improvement over the whole-image baseline, with the Attribute features showing the best performance in most cases. From the object regions we extract localized ImageNet features which also produce some improvement over the whole-image baseline in four out of six cases. 
Since, according to Figure \ref{figure:stat}, color is mentioned in only a tiny fraction of answers to the whole-image questions, we do not include it here.
\smallskip

\begin{table*}[t!]
	\setlength{\tabcolsep}{2pt}
	\footnotesize
    %\hspace{-3mm}
	%\begin{center}   
    \resizebox{\textwidth}{!}{%
 		\begin{tabular}{@{}c|ll||c|c|c||c|c||c|c|c|c@{}} 
            \hline
  {\multirow{2}{*}{Distractor}} & \multicolumn{2}{c||}{\multirow{2}{*}{Question Type}} & \multicolumn{3}{c||}{fc7 Combination } & \multicolumn{2}{c||}{Label Combination} & \multicolumn{4}{c}{CCA Score Combination} \\
            \cline{4-12}
            {\multirow{2}{*}{Type}}& & & Baseline & \multicolumn{2}{c||}{Baseline +} & HICO & HICO + MPII & + Person & + Object & \multicolumn{2}{c}{CCA Ensemble}\\
            && & ImageNet  & \multicolumn{2}{c||}{ Single Best Cue} & + MPII & + Attr. & Score & Score & L2 & L1\\\hline   
            \multirow{10}{*}{Easy} & \multirow{3}{*}{(A)}& {3) Interesting}  & 78.11 & HICO & 79.47 & 79.25 & 79.94 & 80.59 & 80.88 & {\bf82.92} & 82.34\\  
             &       & {4) Past}		    & 79.30 & Attr. & 81.68 & 82.17 & 84.09 & 84.17 & 84.97 & 85.89 & {\bf 85.91}\\ 
             &       & {5) Future}	    & 79.52 & HICO. & 82.29 &82.89 & 84.97 & 84.97 & 85.47 & {\bf86.75} &  86.63\\ 
             \cline{2-12}
             &       \multirow{4}{*}{(B)}& 6) Person's Attribute  & 53.32 & Attr. & 64.50 &59.37 & 68.43 & -- &-- & 68.59 &{\bf 68.68}\\ 
			 &	   &  7) Person's Activity  & 83.89 & HICO & 87.46 &87.23 & 87.26 & -- &-- & 88.11 &{\bf 88.43}\\ 
			 &	   &  8) Person's Location  & 84.59 & Places & 85.70 &84.56 & 84.51 & -- &-- & {\bf86.52} & 86.28\\ 
			 & 	   &  9) Person-Object Relation  & 71.36 & HICO & 75.77 & 75.42 & 75.66 & -- &-- & {\bf 77.77} &77.08\\ 
             \cline{2-12}
             &     \multirow{3}{*}{(C)}& 10) Object's Attribute   & 50.15 & Color & 59.50 &-- &-- &-- &-- & 59.48 &{\bf 59.62}\\
		     &     & 11) Object's Affordance  & 80.56 & Obj.\ VGG & 87.20 &-- &-- &-- &-- & 85.74&{\bf 87.21}\\         
             &     & 12) Object's Position  & 67.79 & Places & 69.41 &-- &-- &-- &-- &69.44 &{\bf 69.71}\\ 
             \cline{2-12}
             &      \multicolumn{2}{c||}{Average}   & 72.86 & & 77.30 &-- &-- &-- &-- &79.12&{\bf 79.19}\\ 
             \hline
             \multirow{10}{*}{Hard} &\multirow{3}{*}{(A)}& {3) Interesting}  & 54.36 & Attr. & 56.02 & 54.11 & 55.37 & 56.25 & 56.31 & {\bf58.37} & 57.92\\ 
             &       & {4) Past}	             & 53.89 & Attr. & 55.90 &55.23 & 58.17 & 58.29 & 59.60 & {\bf61.37} & 61.33\\ 
             &       & {5) Future}	         & 55.19 & Attr. & 57.58 &56.87 & 59.98 & 60.05 & 61.91 & {\bf62.82} & 62.73\\ 
             \cline{2-12}
             &     \multirow{4}{*}{(B)}& 6) Person's Attribute  & 42.55 & Attr. & 54.64 &46.61 & 56.17 & -- &-- & {\bf56.47} & 56.38\\ 
			 &	   &  7) Person's Activity          & 67.56 & HICO & 71.02 &71.35 & 71.42 & -- &-- & 71.00&{\bf 71.68}\\ 
			 &	   &  8) Person's Location        & 64.57 & Places & {\bf 66.66} &62.82 & 62.46 & -- &-- & 66.50 &{\bf 66.66}\\ 
			 &	   &  9) Person-Object Relation & 54.46 & HICO & {\bf 58.72} & 56.68 & 56.88 & -- &-- & 57.80 &57.92 \\ 
             \cline{2-12}
             &     \multirow{3}{*}{(C)}& 10) Object's Attribute   & 44.99 & Color & {\bf 54.73} &-- &-- &-- &-- & {\bf54.75} & 54.73\\
		     &     & 11) Object's	Affordance        & 64.26 & Obj.\ VGG & 67.65 &-- &-- &-- &-- & {\bf67.69} &{\bf 67.69}\\         
             &     & 12) Object's Position          & 56.46 & Places & 57.34 &-- &-- &-- &-- & {\bf 58.22} &58.16\\ 
             \cline{2-12}
             &     \multicolumn{2}{c||}{Average}   & 55.83 & & 60.03 &-- &-- &-- &-- &61.50 &{\bf 61.52}\\ 
             \hline                   
        \end{tabular}
        }
	%\end{center} 
\caption{Results of combining multiple cues. Columns marked ``fc7 Combination'' give key results from Table \ref{table:madlibs_accuracy} for reference. Columns marked ``Label Combination'' show results with combining the class activation vectors from the respective networks. Columns marked ``+ Person Score'' and ``+ Obj. Score'' show the results of a learned combination of the HICO + MPII + Attr. CCA with the region selection scores of Section \ref{subsec:image_region_selection}. 
%\red{
The CCA Ensemble columns shows the results of combining all CCA scores appropriate for each 
question type with weights learned using either the L2 or the L1 regularization. The obtained average results are slightly better in the L1 
case and the weights obtained in this way provide good interpretability (see Table \ref{table:weights}).}
%}
	\label{table:multi_cue}%\vspace{-3mm}
\end{table*}

\noindent{\bf Person Questions.}
For questions about specified people, Table~\ref{table:madlibs_accuracy}(B) reports results with 
features extracted from the provided ground truth person box. Not surprisingly, Attribute features 
give the biggest improvement for Attribute questions, and HICO Action features give the biggest 
improvement for Person's Activity and Person-Object Relationship questions (recall that HICO classes correspond 
to interactions between people and MSCOCO objects). For the latter question type, the ground truth 
object region is also provided; by extracting the ImageNet and Color features from the object box we 
obtain accuracy lower than that of the HICO representation but still higher than that of the whole-image baseline. 
Finally, for Person Location questions, the global Places features work the best. 
This question asks about the place where the person is, \ie the environment around him/her. 
Thus, visual information from the image part outside the person bounding box is more helpful than the
localized information inside the person box which capture more the person appearance rather than 
the appearance of the surrounding location. 
\smallskip

%%%%%%%%%%%%%%%%%%%%

\noindent{\bf Object Questions.}
For questions about specified objects, Table~\ref{table:madlibs_accuracy}(C) reports results with features extracted from the provided ground truth object box. We can see that 
Color features work best for Object's Attribute questions, ImageNet features work best for Object's Affordance questions, and Places features work best for Object's Location questions. 
\smallskip

\label{sec:single_cue_exp}

\subsection{Multi-Cue Results}

\begin{table*}[tb!]
	\setlength{\tabcolsep}{5pt}
	\footnotesize
	\begin{center}
    \resizebox{\textwidth}{!}{%
		\begin{tabular}{c|ll|C{1.5cm}|C{1.5cm}|C{1.5cm}|C{1.5cm}|C{1.5cm}|C{1.5cm}|C{1.5cm}} 
            \hline
            {\multirow{2}{*}{Distractor}} & \multicolumn{2}{c|}{\multirow{3}{*}{Question Type}} & {Full Image} & \multicolumn{3}{c|}{Person Box} & \multicolumn{3}{c}{Object Box}\\ 
            \cline{4-10}
             & & &  B. +        & B. + & HICO + MPII & Person     & B. + & B. +  & Object\\
             Type & & &  Places      & ImageNet  & + Attr.     & Score      & ImageNet  & Color & Score\\\hline
             \multirow{10}{*}{Easy} & \multirow{3}{*}{(A)}&{3) Interesting}  & 0.00 & 0.00  & \cellcolor{red!64}0.64 & 0.00 & \cellcolor{red!36}0.36 & -- & 0.00 \\
             &       &{4) Past}	     & \cellcolor{red!1}0.01 & \cellcolor{red!2}0.02  & \cellcolor{red!63}0.63 & \cellcolor{red!5}0.05 & \cellcolor{red!29}0.29 & -- & 0.00 \\ 
             &        & {5) Future}	 & \cellcolor{red!3}0.03 & \cellcolor{red!1}0.01  & \cellcolor{red!68}0.68 & \cellcolor{red!4}0.04 & \cellcolor{red!24}0.24 & -- & 0.00 \\ 
             \cline{2-10}
             & \multirow{4}{*}{(B)}&   6) Person's  Attribute          & 0.00 & \cellcolor{red!21}0.21  & \cellcolor{red!79}0.79 & -- & -- & -- & -- \\ 
			 &	    &	7) Person's Activity				& \cellcolor{red!8}0.08 & \cellcolor{red!7}0.07  & \cellcolor{red!85}0.85 & -- & -- & -- & -- \\ 				&	   &	8) Person Location				& \cellcolor{red!67}0.67 & 0.00  & \cellcolor{red!34}0.34 & -- & -- & -- & -- \\ 
			 &	    &	9) Person-Object Relation    & \cellcolor{red!11}0.11 & \cellcolor{red!17}0.17  & \cellcolor{red!42}0.42 & -- & \cellcolor{red!23}0.23 & \cellcolor{red!7}0.07 & -- \\ 
             \cline{2-10}
             & \multirow{3}{*}{(C)}& 10) Object's Attribute           & 0.00 & --    & --   & -- & \cellcolor{red!18}0.18 & \cellcolor{red!82}0.82 & -- \\
		     &  & 11) Object's Affordance       & \cellcolor{red!20}0.20 & --    & --   & -- & \cellcolor{red!80}0.80 & 0.00 & -- \\
             &  & 12) Object's Position           & \cellcolor{red!84}0.84 & --    & --   & -- & \cellcolor{red!16}0.16 & 0.00 & -- \\ 
             \hline 
             \multirow{10}{*}{Hard} & \multirow{3}{*}{(A)}&{3) Interesting}  & \cellcolor{red!9}0.09 & \cellcolor{red!15}0.15  & \cellcolor{red!41}0.41 & \cellcolor{red!11}0.11 & \cellcolor{red!23}0.23 & -- & \cellcolor{red!1}0.01 \\
             &        &{4) Past}	            & \cellcolor{red!4}0.04 & \cellcolor{red!5}0.05  & \cellcolor{red!48}0.48 & \cellcolor{red!16}0.16 & \cellcolor{red!17}0.17 & -- & \cellcolor{red!9}0.09 \\ 
             &        & {5) Future}	        & \cellcolor{red!5}0.05 & \cellcolor{red!19}0.19  & \cellcolor{red!31}0.31 & \cellcolor{red!14}0.14 & \cellcolor{red!9}0.09 & -- & \cellcolor{red!22}0.22 \\ 
            \cline{2-10}
		    & \multirow{4}{*}{(B)}&   6) Person's  Attribute    & \cellcolor{red!3}0.03 & \cellcolor{red!35}0.35  & \cellcolor{red!62}0.62 & -- & -- & -- & --       \\ 
 			&	    &	7) Person's Activity			 & 0.00 & \cellcolor{red!48}0.48  & \cellcolor{red!52}0.52 & -- & -- & -- & --  	\\ 				     
            &       &	8) Person's Location			 & \cellcolor{red!100}1.00 & 0.00  & 0.00 & -- & -- & -- & -- 	\\ 
 			&	    &	9) Person-Object Relation & \cellcolor{red!3}0.03 & \cellcolor{red!39}0.39  & \cellcolor{red!13}0.13 & -- & \cellcolor{red!44}0.44 & \cellcolor{red!1}0.01 & --    \\ 
            \cline{2-10}
			& \multirow{3}{*}{(C)}& 10) Object's Attribute      & 0.00 & --    & --   & -- & \cellcolor{red!20}0.20 & \cellcolor{red!80}0.80 & --       \\
	        &       & 11) Object's Affordance       & \cellcolor{red!29}0.29 & --    & --   & -- & \cellcolor{red!71}0.71 & 0.00 & -- \\
			&       & 12) Object's Position           & \cellcolor{red!77}0.77 & --    & --   & -- & \cellcolor{red!23}0.23 & 0.00 & -- \\ 
            \hline           
       \end{tabular}
       }\vspace{-2mm}
	\end{center} 
	\caption{Weights assigned by the CCA score combination %\red{
    (Ensemble L1) %} 
    method to each cue. Questions related to location (types 8, 12) heavily rely on scene predictions, while action and attribute cues (HICO+MPII+Attr. column) are useful for a large variety of question types.}
	\label{table:weights}
    \vspace{-3mm}
\end{table*}

%!TEX root = main.tex
Table~\ref{table:multi_cue} shows the results obtained by integrating multiple cues in a variety of ways. We exclude Scene and Emotion questions from the subsequent analysis: based on Figure \ref{figure:stat}, very few of their answers involve persons and objects, thus, our final cue combination for these question types is simply the concatenation of 
ImageNet and Places as shown in Table \ref{table:madlibs_accuracy}. 

For ease of comparison, the first and second columns of Table~\ref{table:multi_cue} repeat the baseline and highest results from Table~\ref{table:madlibs_accuracy}. The subsequent columns show performance obtained with other cue combinations.
The Label Combination columns of Table~\ref{table:multi_cue} show the results of concatenating the class prediction vectors from the HICO and MPII networks, and from all three person-centric networks (HICO+MPII+Attribute). For HICO+MPII, we observe a small drop in performance over the single best cue on whole-image questions (\ie, in Interesting, Past, Future rows) and location-related questions (Person's Location and Person-Object Relation), probably owing to the reduced feature dimension and loss of global contextual 
information as compared to the 8192-dimensional fc7 combination feature.
On the other hand, HICO+MPII produces results comparable with the best fc7 cue for the Person's Activity question while being much more compact (993 vs. 8192 dimensions). By adding the attribute labels (HICO+MPII+Attribute column), we further improve performance, particularly on the Person's Attribute question.

Recall from Section \ref{subsec:image_region_selection} that for Interestingness, Past, and Future questions, we perform focus region selection and compute Person and Object scores measuring the compatibility of person and object mentions in answers with the selected regions. These scores also provide some useful signal for choosing the correct answer, so we use the procedure of Section \ref{subsec:multi_cue_integration} to learn to combine each of them with the scores from the HICO+MPII+Attribute CCA model. For these two-cue problems, the learning procedure assigns a high weight to the combined action and attribute representation ($w^{HICO+MPII+Attribute}\ge 0.9$) and a small one to the Person and Object scores ($w^{Person/Obj.~Selection}\le 0.1$). 
The resulting accuracies are reported in columns labeled ``+ Person Score'' and ``+ Object Score'' of Table~\ref{table:multi_cue}, and they show small but consistent accuracy improvements over the HICO+MPII+Attribute model, particularly for the hard questions. 

\begin{figure*}[!hp]
\centering
\begin{minipage}{1\textwidth}
\centering
\includegraphics[width=0.95\linewidth]{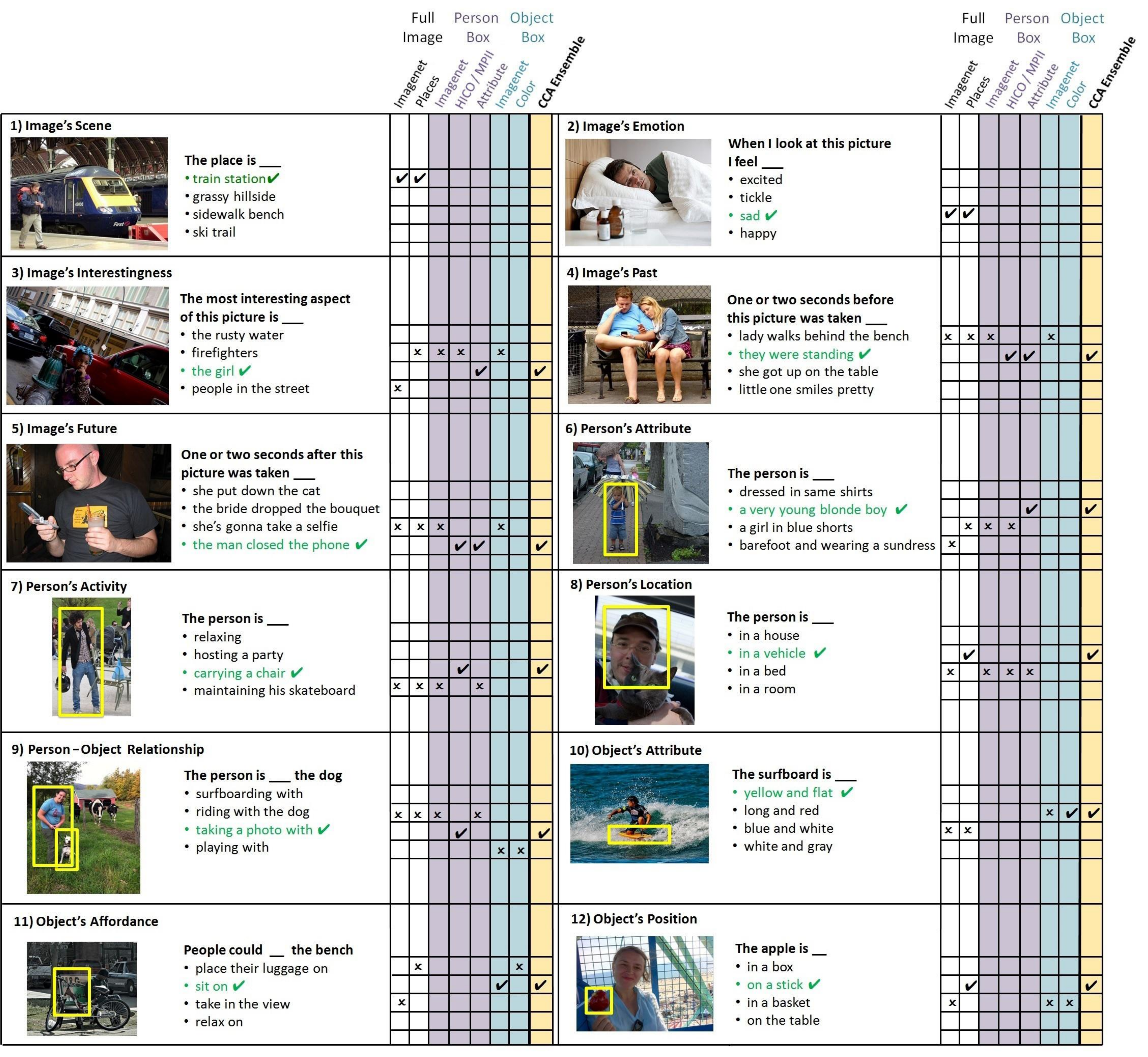}
\caption{Examples of answers selected using each individual cue 
as well as the full ensemble method. The first ImageNet column corresponds to the baseline feature (B.), 
while the following columns correspond to ``B. + X'' features following the same order as in 
Table \ref{table:madlibs_accuracy}. 
Check marks specify that the correct answer has been selected when using the corresponding column
feature for multi-choice answering. The crosses indicate instead a wrong selected answer.}
\label{fig:examples_int_past_fut}
%\vspace{-1mm}
	\end{minipage}
%\end{figure*}
%\begin{figure*}[h]
\begin{minipage}{1\textwidth}
\centering
\includegraphics[width=0.95\linewidth]{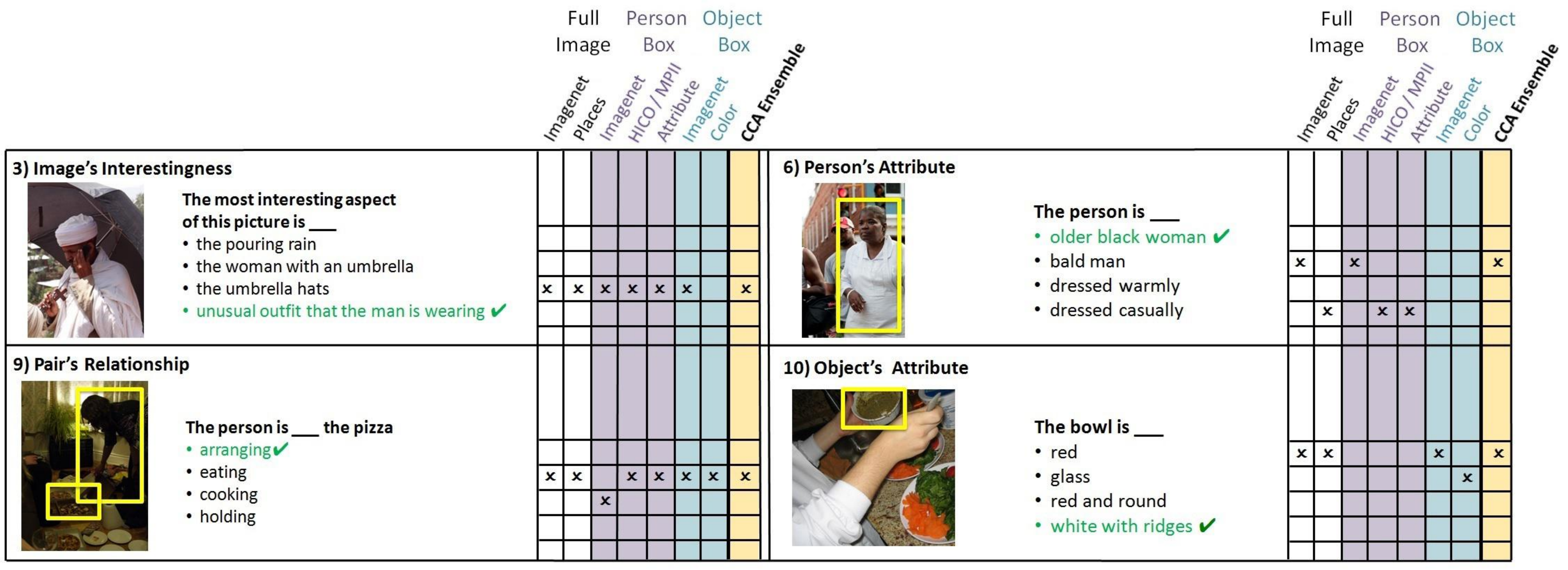}
\caption{Failure cases for four multi-choice question types from the Hard question-answering setting. Examples in the left column involve relatively rare concepts like ``unusual outfit'' and ``arranging the pizza,'' while examples on the right are visually subtle or ambiguous. The crosses indicate a wrong selected answer.}
\label{fig:multicue_failures} 
\end{minipage}
\end{figure*}

The last column of Table \ref{table:multi_cue} gives the performance of the full ensemble score using all the CCA models applicable to a given question type (refer back to Section \ref{subsec:multi_cue_integration} for the list of models). 
%\red{
We report both the results obtained using the L1 regularized weights according to Eq.  (\ref{eq:L1reg}) and its variant based on L2 regularization. The accuracies are similar in both cases, with the L1 case marginally better on average. Using L1 however allows for better understanding the role of each cue:  the per-cue weights for each question type are shown in Table \ref{table:weights}.
%}. 
Generally, the most informative cues for each question type get assigned higher weights (\eg HICO+MPII+Attribute features get high weights for Person's Activity and Person's Attribute questions, but not for Person's Location questions).
From the ``Average'' row of Table \ref{table:multi_cue}, we can observe an improvement of about 1.5\%  
in accuracy with respect to the single best cue and about 6\% with respect to the baseline for both the Easy and
Hard cases.

\begin{table}[tb!]
	
	\begin{center}
    %\resizebox{\textwidth}{!}{%
		\begin{tabular}{@{}c@{~}|@{}c@{}l@{~}|@{~}c@{~}|@{~}c@{}} 
  %\rowfont{\footnotesize} 
  \footnotesize
  {Distr.} & \multicolumn{2}{c|@{~}}{Question} &    CCA &  [Mokarian \\
  Type 		   & \multicolumn{2}{c|@{~}}{Type} & Ensemble		&   et al (2016)]\\
  \hline
  \multirow{11}{*}{Easy} & \multirow{3}{*}{(A)} & 3) Interesting & {\bf 82.34} & 78.20\\
  						 && 4) Past & {\bf 85.91} &80.80\\
                         && 5) Future & {\bf 86.63} & 81.10\\\cline{2-5}
                         &\multirow{4}{*}{(B)}& 6) Person's Attribute& {\bf 68.68}& 56.00\\
                         && 7) Person's Activity& {\bf 88.43} & 83.00\\
                         && 8) Person's Location& {\bf 86.28}& 84.30\\
                         && 9) Person-Object Relation& {\bf 77.08}& 75.30\\\cline{2-5}
                         &\multirow{4}{*}{(C)}& 10) Object's Attribute& 59.62& {\bf 62.40}\\
  						 && 11) Object's Affordance& {\bf 87.21}& 83.30\\
                         && 12) Object's Position& 69.71& {\bf 77.50}\\
                         \cline{2-5}
                         &\multicolumn{2}{@{}c@{}|@{~}}{Average}& {\bf 79.19}& 76.19\\
                         \hline 
\multirow{11}{*}{Hard} &  \multirow{3}{*}{(A)} &3) Interesting & {\bf 57.92} & 54.20\\
  						 && 4) Past & {\bf 61.33} &54.60\\
                         && 5) Future & {\bf 62.73} & 56.10\\\cline{2-5}
                         &\multirow{4}{*}{(B)}& 6) Person's Attribute& {\bf 56.38}& 44.20\\
                         && 7) Person's Activity& {\bf 71.68} & 65.50\\
                         && 8) Person's Location& {\bf 66.66}& 65.20\\
                         && 9) Person-Object Relation& {\bf 57.92}& 55.70\\\cline{2-5}
                         &\multirow{4}{*}{(C)}& 10) Object's Attribute& {\bf54.73}&  45.70\\
  						 && 11) Object's Affordance& {\bf 67.69}& 63.60\\
                         && 12) Object's Position& {\bf58.16}& 56.30\\
                         \cline{2-5}
                         &\multicolumn{2}{@{}c@{}|@{~}}{Average}& {\bf 61.52}& 56.11\\
                         \hline 
        \end{tabular} 
	\end{center} 
	\caption{Comparison of our CCA Ensemble multi cue method against \cite{ashkan16bmvc}.}
	\label{table:mokarian}
    \vspace{-5mm}
\end{table}
To date, the strongest competing system on Visual Madlibs is that of \cite{ashkan16bmvc}.
%\red{
We benchmark our CCA Ensemble method against their results in Table \ref{table:mokarian}
and show that we outperform their approach with an average accuracy improvement of 3 and 5 percentage
points on the easy and hard distractor cases, respectively.
%}
%our average accuracy for the easy and hard distractor cases outperform that of \cite{ashkan16bmvc}
%are 79.19\% and 61.52\% while the results reported in \citep{ashkan16bmvc} are respectively 76.19\% and 56.11\%. 
Our CCA Ensemble results are superior to theirs on every question type except for easy Object Attribute 
%(we get 59.3\% vs. their 62.4\%) 
and Object Location questions.
%(69.2\% vs. 77.4\%). 
For both these questions, we exploit the ground truth object boxes while the method in \citep{ashkan16bmvc}
pool features over multiple regions. It is also relevant to note that in our experiments, we set aside a portion
of the training data for validation while the method in \citep{ashkan16bmvc} exploits nCCA models learned on the 
entire Visual Madlibs training samples.

Finally, Figure \ref{fig:examples_int_past_fut} shows answer choices selected with individual cues for the same questions that were originally shown in Figure \ref{fig:table_overview}, while Figure \ref{fig:multicue_failures} shows a few failure cases.

\label{sec:multi_cue_exp}

\subsection{Learning Shared Embedding Spaces}
\label{sec:shared}
In all the experiments considered so far, we learned a CCA embedding space per question type and per cue. 
However, the questions can be easily grouped on the basis of their main visual focus (whole image, 
persons, and objects) and it is worthwhile to evaluate the performance on the multi-choice question answering task
using shared embedding spaces obtained from each group. This setting allows us to increase the amount of available 
training data for each model while making them more robust to question variability. 

For each cue, we grouped the training data of question types 1-5 on whole image to define a joint
embedding space for group (A), types 6-9 on persons to define a joint embedding space for group (B) 
and types 10-12 on objects to define a joint embedding space for group (C). At test time, 
these models were used to assess the suitability of putative answers by obtaining one set of scores 
for each cue. Finally the cue combination procedure is applied in two ways: either by exploiting the 
new embedding spaces instead of the original ones (group) or by adding the score produced by 
the new embedding spaces to the original ones (combined). In this last case, we actually deal with a doubled 
number of cues. 
The final CCA Ensemble results are collected in Table \ref{table:shared}, where the first column also reports 
as reference the final results of Table \ref{table:multi_cue} obtained with embedding spaces learned on 
separate question types. From the accuracy values, we can conclude that learning shared models 
is beneficial when the question types are quite similar (as in group A) but it is less helpful in
case of higher variability among the question types (group B and C). In particular, among the
question types 10-12, Object's Affordance and Object's Position appear to be the most specific
question types that do not derive any benefit from sharing information amongst each other and with the
Object's Attribute question. The overall effect of question variability
becomes less evident when separate and group model are combined together in the CCA Ensemble.
\begin{table}[t!]
\footnotesize \hspace{-2mm}
\resizebox{\columnwidth}{!}{%
\begin{tabular}{@{}c@{}|@{}l@{~}l@{~}||@{}c@{}|c|@{}c@{}} 
\hline
  {\multirow{1}{*}{Distr.}} & \multicolumn{2}{c||}{\multirow{2}{*}{Question Type}} & \multicolumn{3}{c}{CCA Ensemble} \\
            \cline{4-6}
            {\multirow{1}{*}{Type}}& & &  {\multirow{1}{*}{separate}}       & {\multirow{1}{*}{group}} & combined \\
%                                   & & &                                    &                           &          \\
\hline   
             \multirow{10}{*}{Easy} & \multirow{3}{*}{(A)}& {3) Interesting} &  82.34 & {\bf 82.85} & {\bf 83.40} \\  
              &                                           & {4) Past}		 &  85.91 & {\bf 86.70} &  {\bf 86.36} \\ 
              &                                           & {5) Future}	     & {\bf 86.63} & {\bf 87.42} &  {\bf 87.68} \\ 
              \cline{2-6}
              &       \multirow{4}{*}{(B)}& 6) Per. Attribute  & {\bf 68.68} & 51.38 & 68.46 \\ 
 			 &	   &  7) Per. Activity  & 88.43 & 87.83 & {\bf 88.85} \\ 
 			 &	   &  8) Per. Location  & 86.28 & 84.47 & {\bf 86.76} \\ 
 			 & 	   &  9) Per.-Obj. Relation  &  77.08 & {\bf 77.91} & {\bf 77.97} \\ 
             \cline{2-6}
              &     \multirow{3}{*}{(C)}& 10) Object's Attribute   & 59.62 & 54.91 & {\bf 59.67} \\
		     &     & 11) Obj. Affordance  & {\bf 87.21} & 86.65 & 85.84 \\         
              &     & 12) Obj. Position   &{\bf 69.71} & 64.46 & 64.31 \\ 
              \cline{2-6}
              &      \multicolumn{2}{c||}{Average}   &  79.19 & 76.46 & {\bf 79.93} \\ 
              \hline
             \multirow{10}{*}{Hard} &\multirow{3}{*}{(A)}& {3) Interesting}  &  57.92 & {\bf 58.90} & {\bf 58.17} \\ 
             &       & {4) Past}		    & 61.33 & 58.60 & {\bf 61.86} \\ 
             &       & {5) Future}	        & 62.73 & 62.47 & {\bf 63.42} \\ 
             \cline{2-6}
             &       \multirow{4}{*}{(B)}& 6) Per. Attribute  & 56.38 &  35.96 & {\bf 56.43} \\ 
			 &	   &  7) Per. Activity  & 71.68 & 70.87 & {\bf 72.02} \\ 
			 &	   &  8) Per. Location  & 66.66 & 60.55 & {\bf 66.78} \\ 
			 & 	   &  9) Per.-Obj. Relation  & 57.92 & 56.33 & {\bf 57.97} \\ 
             \cline{2-6}
             &     \multirow{3}{*}{(C)}& 10) Obj. Attribute  & {\bf 54.73} & 50.82 & {\bf 54.73}  \\
		     &     & 11) Obj. Affordance  & {\bf 67.69} & 47.05 & 52.12 \\         
             &     & 12) Obj. Position    & {\bf 58.16} & 53.46 & 53.55 \\ 
             \cline{2-6}
             &      \multicolumn{2}{c||}{Average}    & {\bf 61.52} & 55.50 & 59.71\\              
             \hline                   
\end{tabular}
}
\caption{Results of multiple cue combination obtained with CCA Ensemble when the CCA models are either trained on separate questions or trained on the combination of several question types. 
The first column, separate, reports results from Table~\ref{table:multi_cue}.
The score produced by the shared CCA models can be substituted (group) or 
added (combined) together with those obtained from separate questions. Here, we indicate with bold font all the results that are equal or higher than the corresponding reference 
from separate questions.}
\label{table:shared}\vspace{-4mm}
\end{table}

\subsection{Transferring Learned Embedding Spaces}
\label{sec:cross}

A further test on the robustness of the learned CCA embedding spaces for multiple-choice
question answering can be done by evaluating how transferable they are across several 
question types without additional training. This can be analyzed by testing a CCA model 
on a different question type with respect to that on which it was originally learned. 
We ran extensive experiments on this setting by using the cues that produced the best 
result on the data of each training question and using it on all the other questions 
as test. As expected, the accuracy in this cross-task setting decreases with 
respect to the standard case with training and testing data from the same question type, 
and the performance drop depends on the question similarity. 
This effect is clearly visible in Table \ref{table:transfer} where we provide examples for this setting
which involve whole image questions and on location related question: despite the drop, 
the cross-task recognition rate is still much better than random, indicating a good 
robustness of the models. 
Surprisingly, a model trained on Person Location (type 8) performs better than the standard model on Scene 
(type 1) questions, probably because the trained embedding space learns for a slightly harder task and is more discriminative.

\begin{table}[t!]
\footnotesize 
\resizebox{\columnwidth}{!}{%
\begin{tabular}{@{}c@{}cl@{}c@{}c@{}c@{}} 
\cline{2-6}
                           &      \multicolumn{1}{ |@{}c@{} }{Distr.}                    &   \multicolumn{1}{ |@{}c}{Question}     & \multicolumn{3}{|c|}{Test}\\
\cline{4-6}
                           &      \multicolumn{1}{ |@{}c@{} }{Type}                    &   \multicolumn{1}{ |@{}c}{Type}           & \multicolumn{1}{ |c }{ 3) Interesting} &  4) Past & \multicolumn{1}{ c@{}| }{5) Future}\\    
\hline
\multicolumn{1}{ |@{}c@{}| }{{\multirow{6}{*}{Train}}} & \multicolumn{1}{ @{}c@{}| }{{\multirow{3}{*}{Easy}}}  & 3) Interesting & \multicolumn{1}{ |c }{\bf{79.94}}       & 77.67 & \multicolumn{1}{ c@{}| }{77.39}\\         
\multicolumn{1}{ |@{}c@{}| }{}   						 &  \multicolumn{1}{ @{}c@{}| }{}                        & 4) Past        & \multicolumn{1}{ |c }{79.23}       & \bf{84.09} & \multicolumn{1}{ c@{}| }{82.78}\\   
\multicolumn{1}{ |@{}c@{}| }{}   						 &  \multicolumn{1}{ @{}c@{}| }{}                        & 5) Future      & \multicolumn{1}{ |c }{78.19}       & 83.38 & \multicolumn{1}{ c@{}| }{\bf{84.97}}\\
\cline{2-6}
\multicolumn{1}{ |@{}c@{}| }{}   &\multicolumn{1}{ @{}c@{}| }{{\multirow{3}{*}{Hard}}}    & 3) Interesting & \multicolumn{1}{ |c }{\bf{55.37}}       & 52.68 & \multicolumn{1}{ c@{}| }{52.38}\\         
\multicolumn{1}{ |@{}c@{}| }{}   & \multicolumn{1}{ @{}c@{}| }{}                          & 4) Past        & \multicolumn{1}{ |c }{54.50}       & \bf{58.17} & \multicolumn{1}{ c@{}| }{56.46}\\   
\multicolumn{1}{ |@{}c@{}| }{}   & \multicolumn{1}{ @{}c@{}| }{}                          & 5) Future      & \multicolumn{1}{ |c }{54.23}       & 57.03 & \multicolumn{1}{ c@{}| }{\bf{59.98}}\\
\hline                    
&&&&&\\
\cline{2-6}
                           &     \multicolumn{1}{ |@{}c@{} }{Distr.}       &  \multicolumn{1}{ |@{}c}{Question}            & \multicolumn{3}{|c|}{Test}\\
\cline{4-6}
                           &     \multicolumn{1}{ |@{}c@{} }{Type}         &  \multicolumn{1}{ |@{}c}{Type}                & \multicolumn{1}{ |c }{1) Scene} & 8) Per. Loc. & \multicolumn{1}{ c@{}| }{12) Obj. Pos.}\\    
\hline
\multicolumn{1}{ |@{}c@{}| }{{\multirow{6}{*}{Train}}} & \multicolumn{1}{ @{}c@{}| }{{\multirow{3}{*}{Easy}}}  & 1) Scene & \multicolumn{1}{ |c }{\bf{89.04}}       & 83.68 & \multicolumn{1}{ c@{}| }{52.39}\\         
\multicolumn{1}{ |@{}c@{}| }{}   						 &  \multicolumn{1}{ @{}c@{}| }{}                        & 8) Per. Loc        & \multicolumn{1}{ |c }{\bf{90.14}}       & \bf{85.70} & \multicolumn{1}{ c@{}| }{56.61}\\   
\multicolumn{1}{ |@{}c@{}| }{}   						 &  \multicolumn{1}{ @{}c@{}| }{}                        & 12) Obj. Pos.      & \multicolumn{1}{ |c }{82.76}       & 79.92 & \multicolumn{1}{ c@{}| }{\bf{69.41}}\\
\cline{2-6}
\multicolumn{1}{ |@{}c@{}| }{}   &\multicolumn{1}{  @{}c@{}| }{{\multirow{3}{*}{Hard}}}    & 1) Scene & \multicolumn{1}{ |c }{\bf{73.22}}       & 63.16 & \multicolumn{1}{ c@{}| }{38.25}\\         
\multicolumn{1}{ |@{}c@{}| }{}   & \multicolumn{1}{ @{}c@{}| }{}                          & 8) Per. Loc.        & \multicolumn{1}{ |c }{72.71}       & \bf{66.66} & \multicolumn{1}{ c@{}| }{43.03}\\   
\multicolumn{1}{ |@{}c@{}| }{}   & \multicolumn{1}{ @{}c@{}| }{}                          & 12) Obj. Pos      & \multicolumn{1}{ |c }{59.27}       & 55.46 & \multicolumn{1}{ c@{}| }{\bf{69.41}}\\
\hline               
\end{tabular}
}
\caption{Transfer Learning results obtained by training and testing CCA models on different question types. For the experiments in
the top table we used the combined cue HICO+MPII+Attr., while for the bottom table we used B+Places. Note that when training on the
Person Location question and testing on the Scene question, the obtained performance is higher than training and testing on Scene
for the Easy distractor case.}
\label{table:transfer}
\vspace{-2mm}
\end{table}

\section{Conclusions}
\label{sec:conclusion}
We have shown that features representing different types of image content are helpful for answering multiple choice questions, confirming that external knowledge can be successfully transferred to the this task through the use of deep networks trained on specialized datasets. Further, through the use of an ensemble of CCA models, we have created a system that beats the previous state of the art on the Visual Madlibs dataset.

A detailed analysis of our approach has shown where further work would be beneficial.
Person and object localization may be improved by a better interpretation of the sentences
that does not focus only on separate entities, but understands their relationships and translates 
them into spatial constraints to guide region selection and feature extraction. 
And, of course, training joint image-text models that can better deal with rare and unusual inputs 
remains an important open problem, as exemplified by the questions in the left column of Figure 
\ref{fig:multicue_failures}.

In the future, %\red{
besides testing our approach on other interesting question types currently not covered by the Madlibs dataset (\eg Persons' Emotion, Person-Person Relation),
%} 
we are also interested in extending the study of multi-cue integration strategies to more
open-ended and general VQA tasks that do not rely on pre-specified question templates.  
As done here, we can start from simple feature concatenation to merge 
visual representations for different cues before model learning.
A related idea has been recently exploited in \citep{dualnet} where the concatenated features are
obtained from networks characterized by different architectures but all trained on ImageNet.
This approach can be easily adjusted to use our various domain expert network features
and extend existing VQA methods like those in \citep{askmeanything, CVPR2017WangVQA}.

\noindent
{\bf Acknowledgments.} This material is based upon work supported by the National Science
Foundation under grants 1302438, 1563727, 1405822, 1444234, %\red{
1562098, 1633295, 1452851, %}
Xerox UAC, Microsoft
Research Faculty Fellowship, and the Sloan Foundation Fellowship.

%\begin{acknowledgements}
%\end{acknowledgements}

%BibTeX users please use one of
\bibliographystyle{spbasic}      % basic style, author-year citations
\bibliography{egbib}   % name your BibTeX data base

\end{document}